\documentclass[10pt,twocolumn,letterpaper]{article}

\usepackage[pagenumbers]{paper} 

\usepackage{graphicx}
\usepackage{amsmath}
\usepackage{amssymb}
\usepackage{booktabs}
\usepackage{multirow}
\usepackage{multicol}
\usepackage{lipsum}  
\usepackage{soul}

\usepackage[pagebackref,breaklinks,colorlinks]{hyperref}

\usepackage[capitalize]{cleveref}
\crefname{section}{Sec.}{Secs.}
\Crefname{section}{Section}{Sections}
\Crefname{table}{Table}{Tables}
\crefname{table}{Tab.}{Tabs.}

\begin{document}

\title{TransMVSNet: Global Context-aware Multi-view Stereo Network with Transformers}

\author{Yikang Ding\thanks{Equal Contribution.}\ \ \ Wentao Yuan\footnotemark[1] \ \ Qingtian Zhu \ \ Haotian Zhang \ \ Xiangyue Liu \ \ Yuanjiang Wang\thanks{Project lead.} \ \ Xiao Liu\thanks{Corresponding author (liuxiao@foxmail.com).}\\
 Megvii Technology \\
{\tt\small dyk20@mails.tsinghua.edu.cn, wtyuan@pku.edu.cn, zqt@stu.pku.edu.cn, liuxiangyue@buaa.edu.cn} \\
{\tt\small  \{zhanghaotian, wangyuanjiang, liuxiao\}@megvii.com} \\
\vspace{-0.3em}\\
}

\maketitle

\begin{abstract}
   In this paper, we present TransMVSNet, based on our exploration of feature matching in multi-view stereo (MVS). We analogize MVS back to its nature of a feature matching task and therefore propose a powerful Feature Matching Transformer (FMT) to leverage intra- (self-) and inter- (cross-) attention to aggregate long-range context information within and across images.  
   To facilitate a better adaptation of the FMT, we leverage an Adaptive Receptive Field (ARF) module to ensure a smooth transit in scopes of features and bridge different stages with a feature pathway to pass transformed features and gradients across different scales. 
   In addition, we apply pair-wise feature correlation to measure similarity between features, and adopt ambiguity-reducing focal loss to strengthen the supervision. To the best of our knowledge, TransMVSNet is the first attempt to leverage Transformer into the task of MVS.
   As a result, our method achieves state-of-the-art performance on DTU dataset, Tanks and Temples benchmark, and BlendedMVS dataset.
   The code of our method will be made available at \url{https://github.com/MegviiRobot/TransMVSNet}.\\
\end{abstract}

\section{Introduction}
\label{sec:intro}


Multi-view stereo (MVS) aims to recover the dense 3D presentation with a series of calibrated images, which is an important task of computer vision. Learning-based MVS networks~\cite{yao2018mvsnet,yao2019recurrent,gu2020cascade} have achieved remarkable progress in terms of reconstruction quality and efficiency. Typically, a MVS network extracts image features by a CNN and constructs cost volume via plane sweep algorithm~\cite{collins1996space} in which source images are warped to the reference view. This cost volume is regularized afterwards to estimate the final depth. 

The nature of MVS is a one-to-many feature matching task, in which each pixel of the reference image is supposed to search along the epipolar line in all warped source images and find an optimal depth with the lowest matching cost. Some recent studies~\cite{sun2021loftr,sarlin2020superglue} have proven the importance of long-range global context in feature matching tasks. However, given the aforementioned MVS pipeline, there are two main problems. (a) Local features are well captured by convolutions. The locality of convolved features prevents the perception of global context information, which is essential for robust depth estimation at challenging regions in MVS, \eg poor texture, repetitive patterns, and non-Lambertian surfaces. (b) Besides, when computing matching costs, the features to be compared are simply extracted respectively from each image itself, which is to say, potential inter-image correspondences are not taken into consideration.

\begin{figure}[t]
  \centering
   \includegraphics[width=\linewidth]{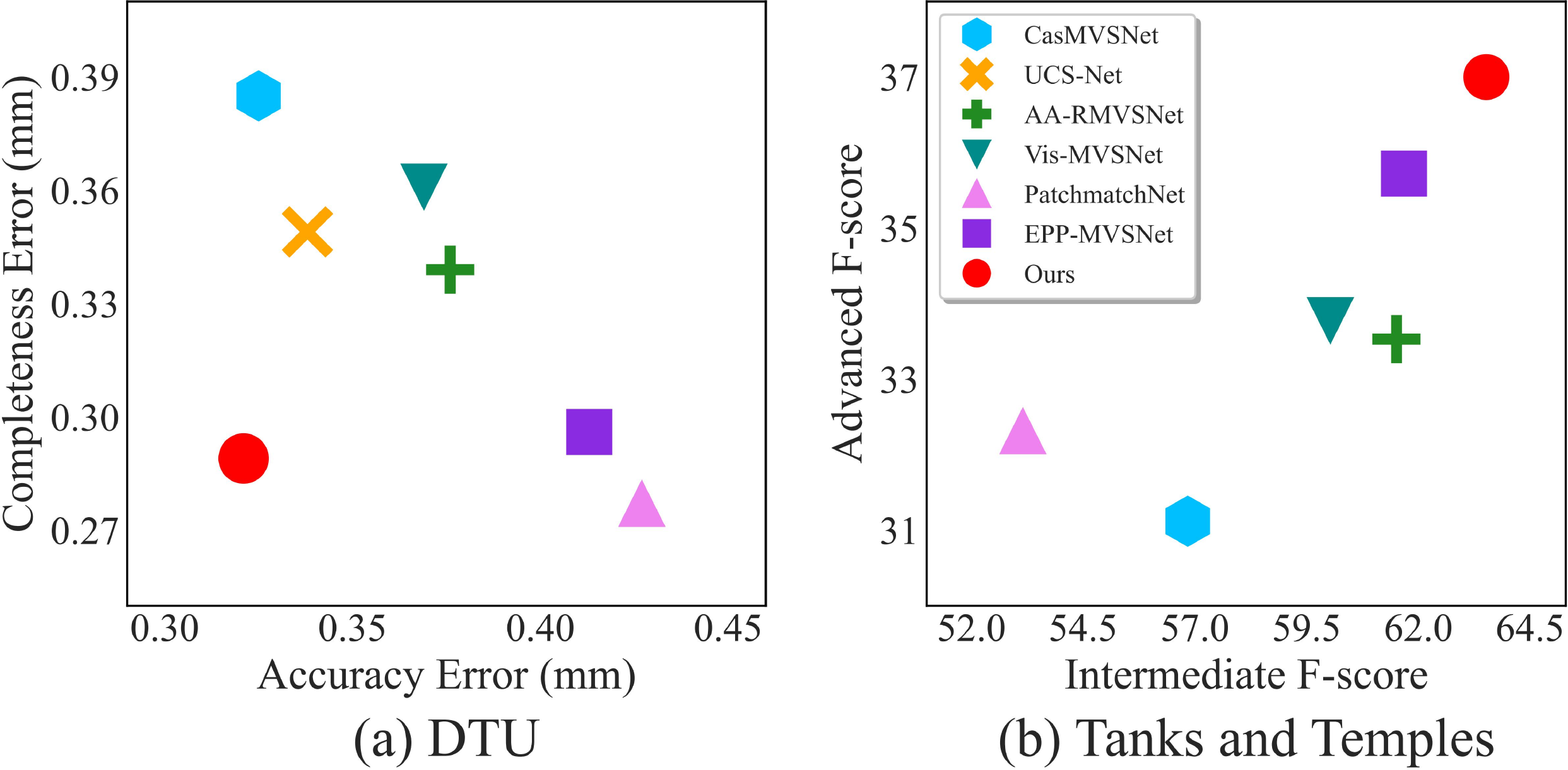}
   \caption{Comparison with state-of-the-art learning-based MVS methods~\cite{gu2020cascade,cheng2020deep,wei2021aa,zhang2020visibility,wang2021patchmatchnet,ma2021epp} on DTU dataset~\cite{aanaes2016large} (\textbf{lower is better}) and Tanks and Temples benchmark~\cite{knapitsch2017tanks} (\textbf{higher is better}).}
   \label{fig:teaser}
\end{figure}
Recently, Transformer~\cite{vaswani2017attention}, which is initially proposed for natural language processing, has drawn considerable attention from the computer vision community for their great performance on vision tasks. Since Transformer utilizes the mechanism of attention and positional encoding for context aggregation, rather than convolutions, it is capable of perceiving global and positionally relevant context information in the true sense. 

To this end, we propose a novel end-to-end deep neural network, namely TransMVSNet, to which a powerful Feature Matching Transformer (FMT) is leveraged to strengthen long-range global context aggregation within and between images. 
To better adapt FMT into an end-to-end learning-based MVS pipeline, we introduce an Adaptive Receptive Field (ARF) module to ensure a smooth transition from locally aggregated features by CNN to features with a global receptive field by FMT. In order to lower runtime memory requirements and train FMT with supervision from high-resolution depth maps, we bridge different scales with a transformed feature pathway. We apply pair-wise feature correlation to measure the similarity between the reference feature map and each of its source feature maps. Afterwards, we follow the coarse-to-fine volume regularization pattern~\cite{gu2020cascade} and adopt focal loss~\cite{lin2017focal}, which better handles samples with ambiguous prediction, to end-to-end train the network.



Thanks to the global context-aware information within and between views, TransMVSNet achieves significant improvement in reconstruction accuracy and completeness simultaneously on DTU dataset~\cite{aanaes2016large} (as shown in \cref{fig:teaser}(a)). Moreover, the overwhelming performance of TransMVSNet can be generalized to more complex scenes, \eg the intermediate and advanced set of Tanks and Templates benchmark~\cite{knapitsch2017tanks} (as shown in \cref{fig:teaser}(b)). To the best of our knowledge, it is the first attempt that takes advantage of Transformer in the task of MVS. Consequently, extensive experiments indicate that our method achieves state-of-the-art performance. We also conduct ablation experiments to demonstrate the effectiveness of each proposed module. Our main contributions are three-fold as follows.

\begin{enumerate}
\setlength{\itemsep}{0pt}
\setlength{\parsep}{0pt}
\setlength{\parskip}{0pt}
    \item[-] We propose a novel end-to-end deep neural network based on a Feature Matching Transformer (FMT), namely TransMVSNet, for robust long-range global context aggregation within and across images.
    \item[-] To better adapt FMT into an end-to-end MVS pipeline, we introduce an ARF module to adaptively adjust the receptive fields of convolved features and apply ambiguity-aware focal loss for training.
    \item[-] Our method achieves state-of-the-art results on DTU dataset, Tanks and Temples benchmark, and BlendedMVS dataset.
\end{enumerate}

\section{Related Work}

\subsection{Learning-based MVS}

In the modern deep era, learning-based methods have been introduced to the task of MVS for better reconstruction accuracy and completeness. MVSNet~\cite{yao2018mvsnet} encodes camera parameters via differentiable homography to build 3D cost volumes, and decouples the MVS task to a per-view depth map estimation task. However, the memory and computation costs are quite expensive due to its 3D U-Net architecture for cost volume regularization.
To alleviate this problem, several networks have been proposed and can be categorized into RNN-based recurrent methods~\cite{yao2019recurrent,yan2020dense,wei2021aa} and coarse-to-fine multi-stage methods~\cite{gu2020cascade,zhang2020visibility,cheng2020deep,yang2020cost}, according to regularization patterns, 
Recurrent methods regularize the 3D cost volumes recurrently, and adopt RNNs to pass features between different depth hypotheses.
Since recurrent methods trade time for space, they are capable of handling images with large resolution but slow in terms of inference speed. Multi-stage methods predict a coarse depth map initially and narrow down the target depth range at a larger resolution based on the previous prediction. Coarse-to-fine methods are able to infer quickly while keeping a relatively small memory consumption. 


Though learning-based MVS methods have achieved promising results, there are still challenging problems remaining, \eg robust estimation at non-Lambertian and low-texture regions or severely occluded areas. 

\subsection{Transformer for Feature Matching}
Transformer~\cite{vaswani2017attention} has been widely used in natural language processing due to its effectiveness and efficiency, and has drawn increasing attention from the computer vision community recently ~\cite{dosovitskiy2021an,carion2020end,ranftl2021vision,pan20213d,liu2021swin}. Considering Transformer's natural superiority to capture global context information by leveraging attention, its ideology has been utilized in the task of feature matching.

SuperGlue~\cite{sarlin2020superglue} utilizes self- and cross-attention in the task of sparse feature matching, leveraging both spatial relationships and visual appearance of the keypoints. SuperGlue achieves impressive performance and becomes the new state of the art.
LoFTR~\cite{sun2021loftr} establishes accurate dense matches with Transformers in a coarse-to-fine manner. By interleaving the self- and cross-attention layers multiple times, LoFTR learns densely arranged and globally consented matching priors in ground-truth matches. 
STTR~\cite{li2021revisiting} models the task of stereo depth estimation from a sequence-to-sequence matching perspective. Transformers with alternating self- and cross-attention along intra- and inter-epipolar line are adopted to capture long-range associations between feature descriptors. 





\begin{figure*}[t]
  \centering
   \includegraphics[width=0.9\linewidth]{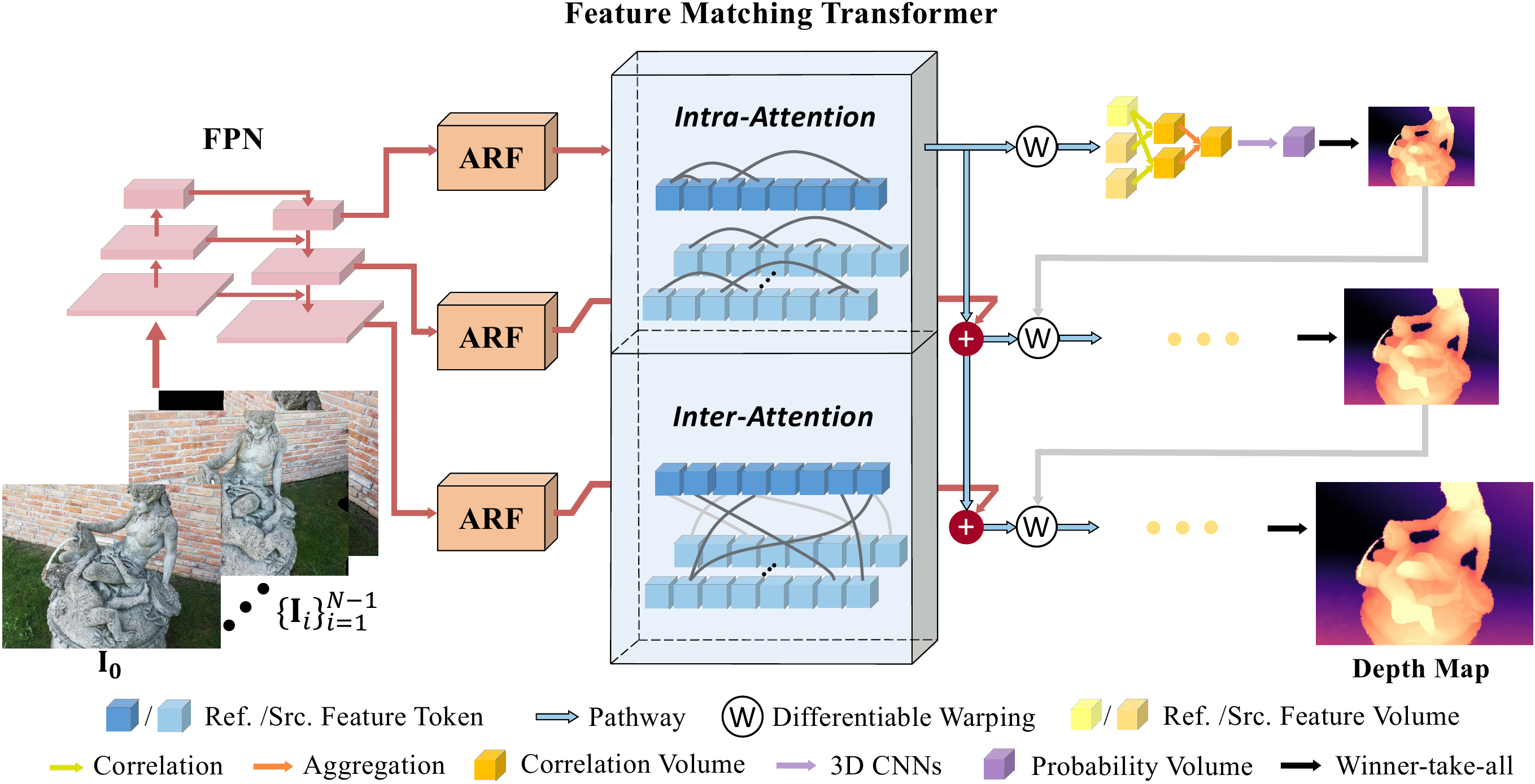}
   \caption{TransMVSNet architecture. TransMVSNet extracts basic features by FPN and introduces ARF modules (\cref{sec:deformable}) to ensure the transit from FPN to Transformer. In the FMT (\cref{sec:transformer}), intra-attention is performed to aggregate global context within images and inter-attention helps feature searching and matching across images. A transformed feature pathway (\cref{sec:pathway}) is connected to pass low-resolution features to higher resolutions and enable back-propagating gradients of all scales to go through FMT. 
   We then apply pixel-wise feature correlation to generate the correlation volumes (\cref{sec:volume}), which are regularized with a coarse-to-fine pattern.}
   \label{fig:network}
\end{figure*}

\section{Methodology}
Given a reference image $\mathbf{I}_0\in \mathbb{R}^{H\times W\times 3} $ and its neighboring images $\{\mathbf{I}_i\}_{i=1}^{N-1}$, as well as their respective camera intrinsics and extrinsics, our method predicts a depth map aligned with $\mathbf{I}_0$. Depth maps of all images are then filtered and fused to obtain the reconstructed dense point cloud.


\subsection{Network Overview}

The overall architecture of our TransMVSNet is illustrated in \cref{fig:network}. TransMVSNet first applies a Feature Pyramid Network (FPN)~\cite{lin2017feature} to extract multi-scale deep image features at three coarse-to-fine levels of resolution. Before handing these features to Transformer, we use the Adaptive Receptive Field (ARF) module, described in \cref{sec:deformable}, to refine the local feature extraction and ensure a smooth transit to Transformer.
To leverage global context information within and between reference and source images, we adopt the Feature Matching Transformer (FMT) to perform intra- and inter-attention. The technical details of FMT are introduced in \cref{sec:transformer}.
To effectively and efficiently propagate transformed features from a low resolution to a higher and make FMT trained with gradients from all scales, we connect all resolutions with a feature pathway described in \cref{sec:pathway}.
To be described in \cref{sec:volume}, for feature maps of $N\times H' \times W' \times F$ processed by FMT, we build a correlation volume of $H' \times W' \times D'\times 1$ for the following regularization by 3D CNNs. $H'$, $W'$ and $F$ denote the height, width and channels of feature maps at current stage, $N$ denotes the number of views and $D'$ denotes the corresponding number of depth hypotheses.   
After obtaining the regularized probability volume, we take the strategy of winner-take-all to determine the final prediction. We apply focal loss with enhanced punishment at ambiguous areas, as described in \cref{sec:loss}, to train TransMVSNet end-to-end.

\subsection{Feature Matching Transformer (FMT)}\label{sec:transformer}

For most cases, learning-based MVS networks construct cost volumes directly from extracted features, ignoring global context information and inter-image feature interaction, which have been proven to be important for improving prediction quality and reducing uncertainty of matching, especially for low-textured regions and repetitive patterns. Aforementioned Transformer-based matching methods handle the problem of feature matching between two views. For MVS, whose nature is a one-to-many matching task, we present a Feature Matching Transformer (FMT), specially customized for MVS. \cref{sec:pre} introduces the preliminaries of attention; \cref{sec:attention} further describes the attention mechanism used in the proposed FMT, especially its customization dedicated to MVS;
\cref{sec:fmt_arch} demonstrates the design of FMT module as a whole.



\subsubsection{Preliminaries}\label{sec:pre}
\paragraph{Scaled dot-product attention}
Analogous to the conventions in information retrieval, features are grouped as query $\mathbf{Q}$, key $\mathbf{K}$ and value $\mathbf{V}$. $\mathbf{Q}$ retrieves relevant information from $\mathbf{V}$ according to the attention weight obtained from the dot product of $\mathbf{Q}$ and $\mathbf{K}$ corresponding to each $\mathbf{V}$. The attention layer is formally denoted as
\begin{equation}
    \mathit{Attention}(\mathbf{Q}, \mathbf{K}, \mathbf{V}) = \mathit{softmax}(\mathbf{Q}\mathbf{K}^\top)\mathbf{V}.
    \label{eq:attention}
\end{equation}
The mechanism of attention measures the feature-wise similarity between $\mathbf{Q}$ and $\mathbf{K}$, and retrieves information from $\mathbf{V}$ according to this computed weight.
Following the practice in \cite{vaswani2017attention}, we adopt multi-head attention, which splits the channel of features into $N_h$ groups (number of heads).

\paragraph{Linear attention}
Multi-head attention~\cite{vaswani2017attention} calculates the attention from the dot product of $\mathbf{Q}$ and $\mathbf{K}$, leading the computation cost growing quadratically with regard to the length of the input sequence. To lower the computation cost, we follow ~\cite{katharopoulos2020transformers} and use Linear Transformer to compute attention. Linear Transformer replaces the original kernel function with
\begin{equation}
    \mathit{Attention}(\mathbf{Q}, \mathbf{K}, \mathbf{V}) = \Phi(\mathbf{Q}) \left(\Phi(\mathbf{K}^\top)\mathbf{V}\right),
    \label{eq:linear attention}
\end{equation}
where $\Phi(\cdot)=elu(\cdot)+1 $ and $elu(\cdot)$ represents the activation function of exponential linear units~\cite{djork2016fast}. Given that the number of channel is far smaller than the length of input sequence, the computation complexity is reduced to linear, making it possible to compute attention upon high-resolution images.

\subsubsection{Intra-attention and Inter-attention}\label{sec:attention}
When $\mathbf{Q}$ and $\mathbf{K}$ vectors are features from the same image, attention layers retrieve relevant information within the given view. This can essentially be seen as intra-image long-range global context aggregation. 
In the other case where $\mathbf{Q}$ and $\mathbf{K}$ vectors are from different views, attention layers then capture cross-relationships across these two views and inter-image feature interaction between images is done in this way. 
In FMT, we perform intra-attention upon both the reference image $\mathbf{I}_0$ and source images $\{\mathbf{I}_i\}_{i=1}^{N-1}$. When computing inter-attention between $\mathbf{I}_0$ and each $\mathbf{I}_i$, only the feature of $\mathbf{I}_i$ is updated.

Here we explain the reason why reference feature $\mathcal{F}_0$ is not supposed to get updated according to source features. When matching the reference image to its neighboring source images, the reference feature should remain invariant to provide an identical target for all source features.
The underlying intuition is that the measurement of similarity is only valid given the same image pair, which indicates that the matching confidence is not comparable universally across different pairs.
We also conduct ablation experiments on this minor topic and get results to support this assumption. Please refer to the Supplementary Material for more information.

\subsubsection{FMT Architecture}\label{sec:fmt_arch}

Different from a typical one-to-one matching task between two views, MVS tackles a one-to-many matching problem, where context information of all views should be considered simultaneously. To this end, we propose the FMT to capture long-range context information within and across images.

\begin{figure}[t]
  \centering
   \includegraphics[width=\linewidth]{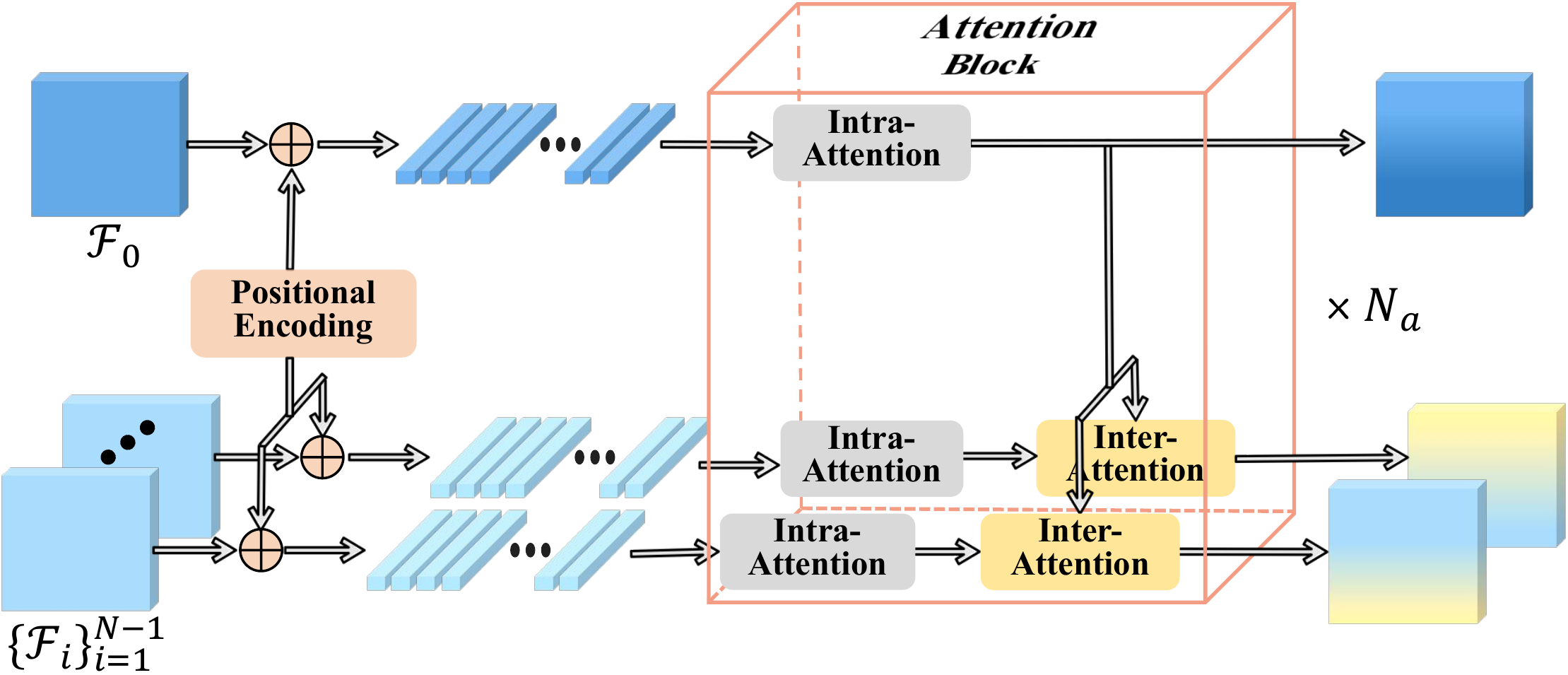}
   \caption{Architecture of the Feature Matching Transformer. FMT performs positional encoding to all features maps and flattens them at the spatial dimension. Then the attention blocks get involved and perform intra- and inter-attention upon features. Note that the number of attention blocks $N_a$ is set as 4 in our implementation.}
   \label{fig:FMT-arch}
\end{figure}

The architecture of FMT is illustrated in \cref{fig:FMT-arch}. We follow \cite{sarlin2020superglue} and add positional encoding, which implicitly enhances positional consistency and makes FMT robust to feature maps with different resolutions.
Each view's corresponding flattened feature map $\mathcal{F}\in \mathbb{R}^{H'W'\times F}$ is processed by $N_a$ attention blocks sequentially. 
Within each attention block (see \cref{fig:FMT-arch}), the reference feature $\mathcal{F}_0$ and each source feature $\mathcal{F}_i$ firstly compute intra-attention with shared weights, where all features are updated with their respective embedded global context information.
Afterwards, the unidirectional inter-attention is performed, with which $\mathcal{F}_i$ is updated according to retrieved information from $\mathcal{F}_0$.


\subsection{Transformed Feature Pathway}\label{sec:pathway}

The Transformer we leverage only performs on feature maps at a rather low resolution since both learning-based MVS and Transformer acquire a massive amount of memory and computation. It remains a problem that how to effectively pass the transformed features from a low resolution to a higher. Besides, we expect the FMT to be trained with supervision from all image scales. We therefore design a transformed feature pathway to fulfill this job. As shown in \cref{fig:network}, feature maps processed by FMT are interpolated to a higher resolution and added to the corresponding raw feature maps at the next image scale.

\subsection{Adaptive Receptive Field (ARF) Module}\label{sec:deformable}

Transformer implicitly encodes global context information into feature maps via positional encoding, which we can roughly perceive as convolution layers with a global receptive field. On the contrary, FPN~\cite{lin2017feature}, which is adopted as the basic feature extractor of the proposed network, mainly focuses on the context within a relatively local neighborhood. There is apparently a gap between these two modules in terms of context ranges, which is detrimental to both feature forwarding and end-to-end training.

To this end, we insert an Adaptive Receptive Field Module between FPN and FMT, to adaptively adjust the scope of extracted features. The ARF module is implemented by deformable convolution~\cite{dai2017deformable,zhu2019deformable}, which learns extra offsets for sampling position and is able to adaptively enlarge the receptive fields according to the local context. 

\subsection{Correlation Volume Construction}\label{sec:volume}

 We apply differentiable warping to align all images to the reference view. The warping between a pixel $\mathbf{p}$ at the reference view and its corresponding pixel $\hat{\mathbf{p}}$ at the source view under depth hypothesis $d$ is defined as
\begin{equation}
    \hat{\mathbf{p}} = \mathbf{K}[\mathbf{R}(\mathbf{K}_0^{-1}\mathbf{p}d)+\mathbf{t}],
\end{equation}where $\mathbf{R}$ and $\mathbf{t}$ denote the rotation and translation between the two views. $\mathbf{K}_0$ and $\mathbf{K}$ are the intrinsic matrices of the reference and source camera. The warped feature maps are bilinearly interpolated to remain the original resolution. By discretizing the known depth space into $D$ depth values, we are able to classify each pixel as one of these values. 

Pair-wise feature correlation at position $\mathbf{p}$ is 
\begin{equation}
    c^{(d)}_i(\mathbf{p}) =<\mathcal{F}_0(\mathbf{p}),\hat{\mathcal{F}}^{(d)}_i(\mathbf{p})>,
    \label{eq:correlation}
\end{equation}
where $\hat{\mathcal{F}}^{(d)}_i$ denotes the warped $i$-th source feature map at depth $d$. In this way, the channel number is reduced to 1, alleviating subsequent memory consumption at regularization. To aggregate all $N-1$ pair-wise correlation volumes, we consider that each pixel in the height and width dimension of 3D correlation volume has different saliency but is consistent in the depth dimension. We therefore assign a pixel-wise weight map with its maximum correlation in the depth dimension. The aggregated correlation volume is then defined as
\begin{equation}
    C^{(d)}(\mathbf{p}) = \sum_{i=1}^{N-1} \max_{d}\{c_i^{(d)}(\mathbf{p})\}\cdot c_i^{(d)}(\mathbf{p}).
\end{equation}

\subsection{Loss Function}\label{sec:loss}

\begin{figure}[t]
  \centering
   \includegraphics[width=\linewidth]{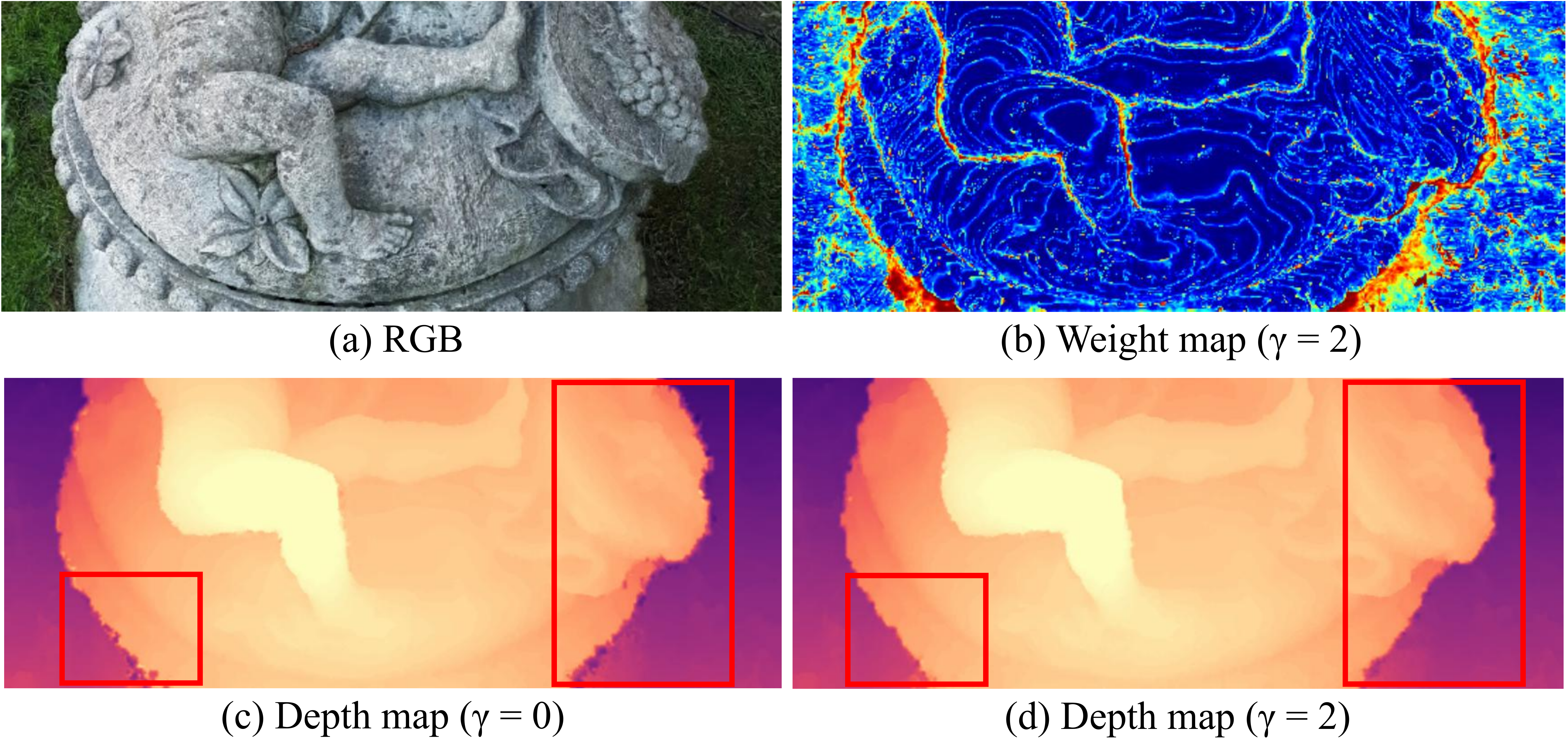}
   \caption{Results visualization of focal loss. (a) Raw image. (b) Focal weight map $(1-P)^\gamma$ when the $\gamma=2$. (c) Depth map when the network is trained with $\gamma=0$. (d) Depth map when the network is trained with $\gamma=2$. Focal loss focuses on pixels with low prediction probability, which normally appear in boundary regions.}
   \label{fig:focal}
\end{figure}

Previous coarse-to-fine attempts~\cite{gu2020cascade,zhang2020visibility,yang2020cost} mainly adopt $\ell$1-based depth regression loss that supervises the absolute distance between prediction and ground truth. We instead apply focal loss~\cite{lin2017focal} that treats depth estimation as a classification task to strengthen the one-hot supervision at ambiguous areas. The focal loss at each depth estimation stage is
\begin{equation}
    \mathcal{L}=\sum_{\mathbf{p}\in \{\mathbf{p}_{v}\}}-(1-P^{(\tilde{d})}(\mathbf{p}))^\gamma \log\left(P^{(\tilde{d})}(\mathbf{p})\right),
\end{equation}
where $P^{(d)}(\mathbf{p})$ denotes predicted probability of depth hypothesis $d$ at pixel $\mathbf{p}$ and $\tilde{d}$ represents the depth value closest to the ground truth among all hypotheses. $\{\mathbf{p}_{v}\}$ represents a subset of pixels with valid ground truth. Specially, focal loss degrades to cross entropy loss  when the focusing parameter $\gamma$ equals 0. Empirically, $\gamma=2$ fits more complicated scenarios and $\gamma=0$ can produce good enough results for relatively simple scenarios. \cref{fig:focal} shows the effect of focal loss on boundary regions, where focal loss helps to estimate more accurate boundary than cross entropy loss.

\section{Experiments}
\subsection{Datasets}
DTU~\cite{aanaes2016large} is captured under well-controlled laboratory conditions with a fixed camera trajectory and contains 128 scans with 49 views under 7 different lighting conditions. Following the setting of MVSNet~\cite{yao2018mvsnet}, we split the dataset into 79 training scans, 18 validation scans, and 22 evaluation scans.
BlendedMVS dataset~\cite{yao2020blendedmvs} is a large-scale synthetic dataset for multi-view stereo training and contains a variety of objects and scenes. The dataset is split into 106 training scans and 7 validation scans.
Tanks and Temples~\cite{knapitsch2017tanks} is a public benchmark acquired in realistic conditions. It contains an intermediate subset of 8 scenes and an advanced subset of 6. Different scenes have different scales, surface reflection, and exposure conditions.

\subsection{Implementation Details}
\begin{figure*}[t]
  \centering
   \includegraphics[width=\linewidth]{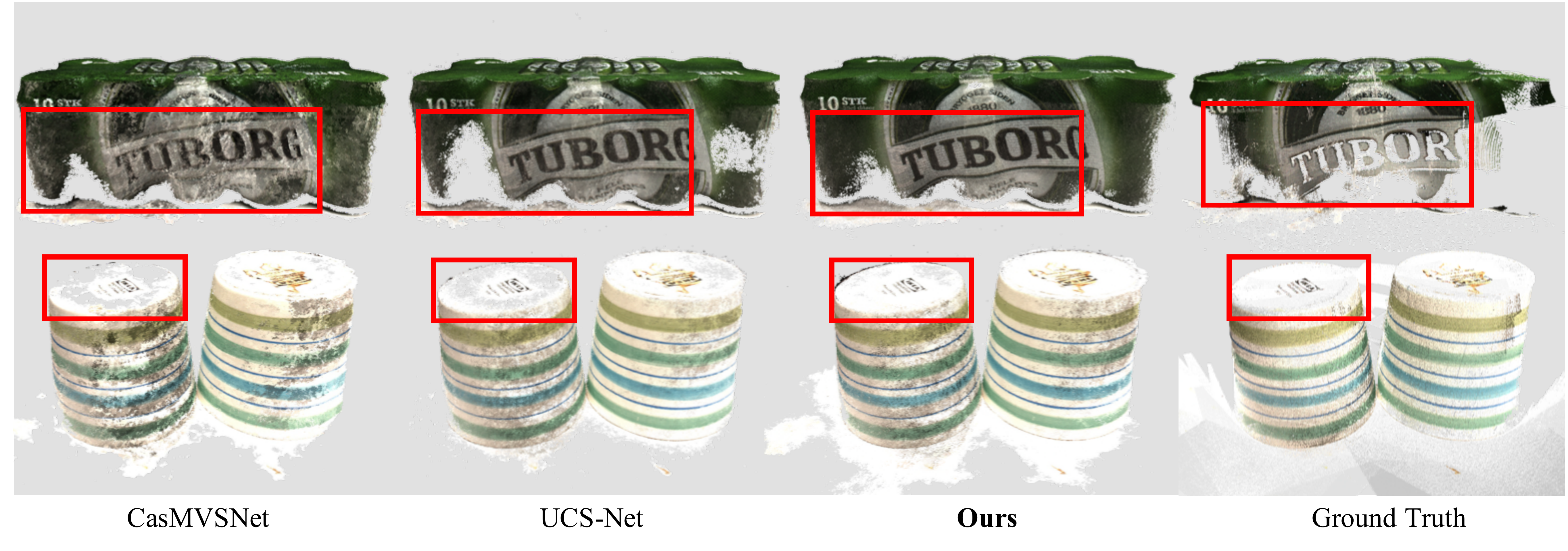}
   \caption{Comparison of reconstructed results with state-of-the-art coarse-to-fine methods~\cite{gu2020cascade,cheng2020deep} on DTU evaluation set~\cite{aanaes2016large}.}
   \label{fig:dtu_compare}
\end{figure*}

We implement TransMVSNet with PyTorch and train it on DTU training set~\cite{aanaes2016large}. 
At training phase, we set the number of input images $N=5$ and image resolution as $512\times 640$. For coarse-to-fine regularization, depth hypotheses are sampled from $425mm$ to $935mm$; the number of plane sweeping depth hypotheses of each stage is respectively 48, 32, and 8; the corresponding depth interval decays by 0.25 and 0.5 from the coarsest stage to the finest stage.
The model is trained with Adam for 10 epochs with an initial learning rate of 0.001, which decays by a factor of 0.5 respectively after 6, 8, and 12 epochs. We set $\gamma=0$ for training on DTU.
The batch size is 1 on 8 NVIDIA RTX 2080Ti GPUs and in total, the training phase takes about 16 hours and occupies 10\textit{GB} memory of each GPU.



For depth filtering and fusion, we follow the dynamic checking strategy proposed in \cite{yan2020dense}, in which both confidence thresholding and geometric consistency are applied. 


\subsection{Experimental Performance}


\paragraph{Evaluation on DTU dataset}
We evaluate the proposed method on the evaluation set of DTU dataset~\cite{aanaes2016large} with official evaluation metrics. 
We set $N = 5$ and the input resolution as $864 \times 1152$ at evaluation phase. 
As is visualized in \cref{fig:dtu_compare}, benefiting from the mechanism of  intra- and inter-attention in FMT, TransMVSNet is able to yield denser and complete point clouds with more details preserved.  
Quantitative comparisons are shown in \cref{tab:dtu}. Accuracy and Completeness are the two official metrics. Accuracy measures the mean absolute point-cloud-to-point-cloud distance from the MVS reconstruction to ground truth, while Completeness measures the opposite. 
The Overall is the average of Accuracy and Completeness, which indicates the overall performance of models.
TransMVSNet achieves competitive performance in Accuracy and Completeness and outperforms all known methods in Overall by a large margin.

\begin{table}[t]
    \centering
    \resizebox{\linewidth}{!}{
    \begin{tabular}{lccc}
    \hline
    \textbf{Method} & Acc.($mm$) & Comp.($mm$) & Overall($mm$)\\
    \hline
    Gipuma~\cite{galliani2015massively} & \textbf{0.283} & 0.873 & 0.578 \\
    COLMAP~\cite{schonberger2016mvs} & 0.400 & 0.664 & 0.532 \\
    \hline
    R-MVSNet~\cite{yao2019recurrent} & 0.385 & 0.459 & 0.422 \\ 
    $D^2$HC-RMVSNet~\cite{yan2020dense} & 0.395 & 0.378 & 0.386 \\
    AA-RMVSNet~\cite{wei2021aa} & 0.376 & 0.339 & 0.357 \\
    \hline
    Vis-MVSNet~\cite{zhang2020visibility} & 0.369  & 0.361 & 0.365 \\
    CasMVSNet~\cite{gu2020cascade} & 0.325 &  0.385 & 0.355 \\ 
    UCS-Net~\cite{cheng2020deep} & 0.338 & 0.349 & \underline{0.344} \\
    PatchmatchNet~\cite{wang2021patchmatchnet} & 0.427 & \textbf{0.277} & 0.352 \\
    EPP-MVSNet~\cite{ma2021epp} & 0.413 & 0.296 & 0.355 \\
    \hline
    \rule{0pt}{9pt}\
    \textbf{TransMVSNet} &  \underline{0.321} & \underline{0.289} & \textbf{0.305} \\[2pt]
   \hline
    \end{tabular}}
    \caption{Quantitative results on DTU evaluation set~\cite{aanaes2016large} (\textbf{lower is better}). \textbf{Bold} figures indicate the best and \underline{underlined} figures indicate the second best. Compared to non-learning methods, RNN-based methods and coarse-to-fine methods, TransMVSNet outperforms all known methods by a large margin.}
    \label{tab:dtu}
\end{table}

\begin{table*}[t]
    \centering
    \setlength\tabcolsep{1.8pt}
    \resizebox{\linewidth}{!}{
    \begin{tabular}{l|c|cccccccc|c|cccccc}
    \hline
    \hline
    \textbf{Method} &  \textbf{Int.Mean} & Family & Francis & Horse & L.H. & M60 & Panther & P.G. & Train & \textbf{Adv.Mean} & Auditorium & Ballroom & Courtroom & Museum & Palace & Temple \\
    \hline
    COLMAP~\cite{schonberger2016mvs} & 42.14 & 50.41 & 22.25 & 26.63 & 56.43 & 44.83 & 46.97 & 48.53 & 42.04 & 27.24 & 16.02 & 25.23 & 34.70 & 41.51 & 18.05 & 27.94\\
    ACMM~\cite{xu2019multi}&57.27	&	69.24	&51.45	&46.97	&63.20&	55.07&	57.64&	60.08&	54.48 & 34.02 & 23.41&32.91& \underline{41.17} &48.13&23.87&34.60\\
    \hline
    DeepC-MVS~\cite{kuhn2020deepc} &	59.79&	71.91&	54.08&	42.29&	 \textbf{66.54} &55.77& \textbf{67.47} &	60.47& \textbf{59.83} & 34.54 & \textbf{26.30} & 34.66 & \textbf{43.50} & 45.66 & 23.09 & 34.00\\
    AttMVS~\cite{luo2020attention} &	{60.05} &	73.90 &	 \underline{62.58} &	44.08 &	\underline{64.88} &	56.08&	59.39 &	 \textbf{63.42} & 56.06 & 31.93 & 15.96 &27.71& 37.99 & \textbf{52.01} & 29.07 & 28.84\\
    \hline
    CasMVSNet~\cite{gu2020cascade} & 56.84	&	76.37&	58.45&	46.26&	55.81&	56.11&	54.06&	58.18&	49.51 & 31.12 & 19.81 & 38.46 & 29.10 & 43.87 & 27.36 & 28.11\\
    Vis-MVSNet~\cite{zhang2020visibility}& 60.03 &	77.40&	60.23&	47.07&	63.44&	62.21&	57.28&	{60.54}&	52.07 & 33.78 & 20.79 &	38.77 &	32.45 & 44.20 & 28.73 & 37.70\\
    PatchmatchNet~\cite{wang2021patchmatchnet} & 53.15 & 66.99 & 52.64 & 43.24 & 54.87 & 52.87 & 49.54 & 54.21 & 50.81 & 32.31 & 23.69 & 37.73 & 30.04 & 41.80 & 28.31 & 32.29\\
    EPP-MVSNet~\cite{ma2021epp} & \underline{61.68} & \underline{77.86} & 60.54 & \underline{52.96} & 62.33 & 61.69 & \underline{60.34} & \underline{62.44} & 55.30 & \underline{35.72} & 21.28 & 39.74 & 35.34 & \underline{49.21} & \underline{30.00} & \textbf{38.75}\\
    \hline
    R-MVSNet~\cite{yao2019recurrent} & 50.55 & 73.01 & 54.46 & 43.42 & 43.88 & 46.80 & 46.69 & 50.87 & 45.25 & 29.55 & 19.49 & 31.45 & 29.99 & 42.31 & 22.94 & 31.10\\
    AA-RMVSNet~\cite{wei2021aa} &	61.51 & 77.77 &	59.53&	51.53 &	64.02&	\textbf{64.05} & 59.47&	60.85&	54.90 & 33.53&20.96 & \underline{40.15} & 32.05 & 46.01 & 29.28 & 32.71\\
    \hline
    \rule{0pt}{12pt} 
    \textbf{TransMVSNet} & \textbf{63.52} & \textbf{80.92} & \textbf{65.83} & \textbf{56.94} & 62.54 & \underline{63.06} & 60.00 & 60.20 & \underline{58.67} & \textbf{37.00} & \underline{24.84} & \textbf{44.59} & 34.77 & 46.49 & \textbf{34.69} & \underline{36.62}\\[3pt]
    \hline
    \hline
    \end{tabular}}
    \caption{Benchmarking results on the Tanks and Temples~\cite{knapitsch2017tanks}. The evaluation metric is mean F-score (\textbf{higher is better}). \textbf{Bold} figures indicate the best and \underline{underlined} figures indicate the second best. TransMVSNet achieves state-of-the-art performance on both the intermediate and the advanced leaderboards of Tanks and Temples benchmark (Nov. 12, 2021).}
    \label{tab:tnt}
\end{table*}

\begin{figure*}[t]
  \centering
   \includegraphics[width=\linewidth]{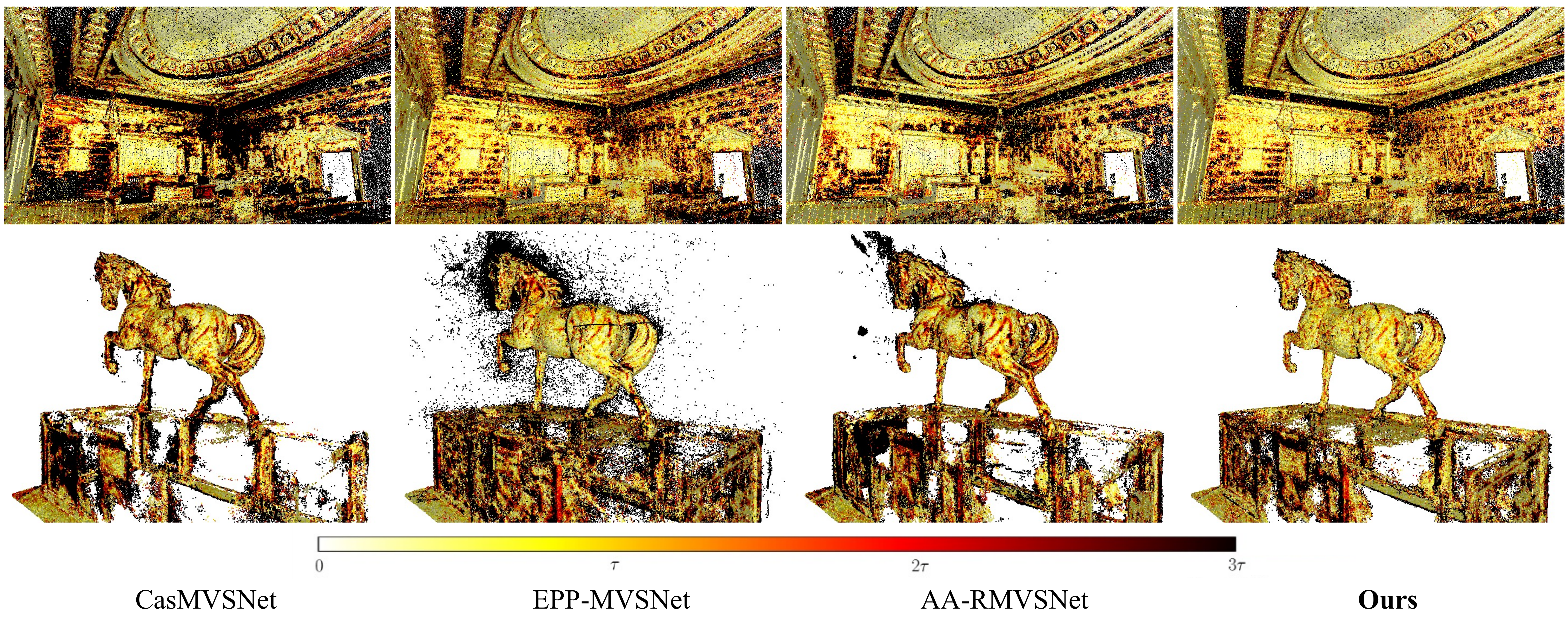}
   \caption{Comparison of reconstructed results with several state-of-the-art methods~\cite{gu2020cascade,cheng2020deep,wei2021aa} on Tanks and Temples benchmark~\cite{knapitsch2017tanks}. $\tau$ is the scene-relevant distance threshold determined officially and darker regions indicate larger error encountered with regard to $\tau$. The first row shows Recall on the scene of Courtroom ($\tau=10mm$); the second row shows Precision on the scene of Horse ($\tau=3mm$).}
   \label{fig:tnt_compare}
\end{figure*}

\begin{figure*}[t]
  \centering
   \includegraphics[width=\linewidth]{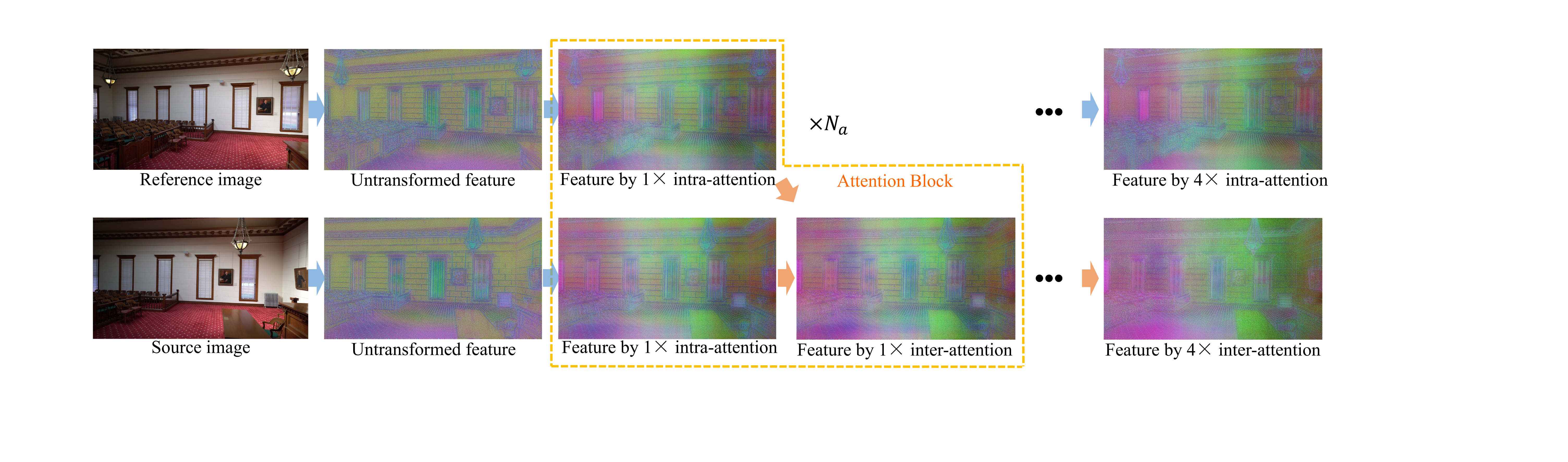}
   \caption{Evolution of feature maps via FMT on the scene Courtroom of Tanks and Temples benchmark~\cite{knapitsch2017tanks}. We apply PCA to reduce the number of feature channels to 3 and visualize the results with RGB. Images of the first row show features of the reference view, which are only updated by intra-attention in FMT; images of the second row represent features of a source view, which get updated by both intra- and inter-attention layers.}
   \label{fig:feature_map}
\end{figure*}

\paragraph{Benchmarking on Tanks and Temples}
To demonstrate the generalization ability of our method, we test our method on Tanks and Temples benchmark~\cite{knapitsch2017tanks}. 
To boost the performance on real-world scenes, we fine-tune TransMVSNet on the training set of the BlendedMVS dataset~\cite{yao2020blendedmvs} using the original image resolution ($ 576\times 768 $), $N = 5$ and $\gamma=2$. 

For evaluation on Tanks and Temples, the camera parameters, depth ranges, and neighboring view selection are aligned with R-MVSNet~\cite{yao2019recurrent}. We use images of the original resolution for inference. Quantitative comparisons on Tanks and Temples are shown in \cref{tab:tnt} and the metrics are mean F-score.
TransMVSNet outperforms all existing learning-based MVS methods on both leaderboards, demonstrating the effectiveness and generalizability of our method. \cref{fig:tnt_compare} shows qualitative results on the scene Courtroom of advanced set and Horse of intermediate set. TransMVSNet yields more reliable points at low-textured areas and sophisticated surfaces. Specially, we visualize the process of feature evolution of a pair of views in \cref{fig:feature_map}. In such a typically challenging scene with poor texture and repetitive patterns, FMT manages to capture position-dependent features and aggregate global context within and across different views.

\paragraph{Evaluation on BlendedMVS dataset}
Both DTU~\cite{aanaes2016large} and Tanks and Temples~\cite{knapitsch2017tanks} apply evaluation metrics towards point clouds. We further demonstrate the quality of depth maps, which are the direct outputs by TransMVSNet, on BlendedMVS validation dataset~\cite{yao2020blendedmvs}. We set $N=5$ and image resolution as $512\times 640$, and apply the evaluation metrics described in \cite{darmon2021deep} where depth values are normalized to make depth maps with different depth ranges comparable.

Some quantitative results are illustrated in \cref{tab:bld}. EPE stands for the endpoint error, which is the average $\ell$-1 distance between the prediction and the ground truth depth; $e_1$ and $e_3$ represent the proportion in $\%$ of pixels with depth error larger than 1 and larger than 3.
Compared with other methods, TransMVSNet achieves impressive results, demonstrating its capability of yielding high-quality depth maps. Please refer to the Supplementary Material for more point cloud results.

\begin{table}[t]
    \centering
    \footnotesize
    \setlength{\tabcolsep}{4mm}
    \begin{tabular}{lccc}
    \hline
    \textbf{Method} & EPE & $e_1$ & $e_3$ \\
    \hline
    MVSNet~\cite{yao2018mvsnet} & 1.49 & 21.98 & 8.32 \\
    CVP-MVSNet~\cite{yang2020cost} & 1.90 &  19.73 & 10.24 \\
    CasMVSNet~\cite{gu2020cascade} & 1.43 &  19.01 & 9.77 \\ 
    Vis-MVSNet~\cite{zhang2020visibility} & 1.47  & 15.14 & 5.13 \\
    EPP-MVSNet~\cite{ma2021epp} & 1.17 & 12.66 & 6.20 \\
    \hline
    \rule{0pt}{9pt}
    \textbf{TransMVSNet} & \textbf{0.73}  & \textbf{8.32} & \textbf{3.62} \\[2pt]
   \hline
    \end{tabular}
    \caption{Quantitative results towards predicted depth maps on BlendedMVS validation set~\cite{yao2020blendedmvs} (\textbf{lower is better}). }
    \label{tab:bld}
\end{table}


\subsection{Ablation Study}
We perform ablation studies to analyze the effectiveness and costs of different modules. The implemented baseline is basically based on CasMVSNet~\cite{gu2020cascade}, which applies feature correlation and is trained with $\ell$-1 loss. 
All the experiments are performed with the same hyperparameters. 

As shown in \cref{tab:ablation}, after applying focal loss, the overall performance improves by 1.7\% while the computational costs remain unchanged. Due to the computational efficiency of Linear Transformer, we are able to leverage FMT with little additional costs in terms of memory and MACs but its inference speed is nearly 1.4 times slower. With the transformed feature pathway, both Completeness and Overall performance get boosted while there is almost no increase in its memory occupancy, indicating the effectiveness and efficiency of the pathway. With ARF module attached, the full TransMVSNet is able to achieve state-of-the-art performance by a large margin. ARF module brings considerable computational costs in all aspects. After all, the inference time is still within one second, which is acceptable compared to RNN-based methods~\cite{yao2019recurrent,yan2020dense,wei2021aa}.


\begin{table}[t]
    \centering
    \setlength\tabcolsep{2pt}
    \resizebox{\linewidth}{!}{
    \begin{tabular}{ccccc|ccc|ccc}
    \hline
      \multicolumn{1}{}{aaa} & \multicolumn{4}{c|}{\textbf{Model Settings}}  & \multicolumn{3}{c|}{\textbf{Mean Distance}} & \multirow{2}{*}{Mem.} & \multirow{2}{*}{MACs} &  \multirow{2}{*}{Time}\\ 
      & F.L. & FMT & Pathway & ARF & Acc.  & Comp. & Overall & & & \\
       \hline
       (a) & & &  &  &  0.351 & 0.339 & 0.345 & 3244 & 212 & 0.271 \\
       (b) & \checkmark&  &  &  &  0.343 & 0.335 & 0.339 & 3244 & 212 & 0.271 \\
       (c) & \checkmark& \checkmark&   &  & 0.335 & 0.310 & 0.323 & 3288 & 235 & 0.638 \\
       (d) & \checkmark& \checkmark&  \checkmark&  &  0.332 & 0.298 & 0.315 & 3288 & 241 & 0.677 \\
       (e) & \checkmark& \checkmark & \checkmark  &  \checkmark & \textbf{0.321} & \textbf{0.289} & \textbf{0.305} & 3778 & 435 &0.996 \\
    \hline
    \end{tabular}}
    \caption{Quantitative performance with different components on DTU evaluation dataset~\cite{aanaes2016large}. F.L. is short for focal loss.
    The unit is \textit{MB} for memory occupancy (Mem.), \textit{G} for multiply–accumulate operations (MACs) and \textit{second} for inference time.}
    \label{tab:ablation}
\end{table}

\section{Discussions}
\subsection{Comparisons to Related Work}\label{sec:compare}
\paragraph{TransMVSNet vs. CasMVSNet} Our architecture is based on the coarse-to-fine regularization pattern proposed by CasMVSNet~\cite{gu2020cascade}. The main difference is that we introduce Transformer to capture long-range global context for better feature matching over multiple views. Using the coarse-to-fine manner brings more computation efficiency while remarkable performance is also achieved. 

\paragraph{TransMVSNet vs. LoFTR} LoFTR~\cite{sun2021loftr} interleaves self- and cross-attention layers multiple times along flattened feature maps to estimate dense matching between a pair of images. Different from one-to-one matching tasks, MVS is actually a one-to-many matching task. We thus propose the FMT module to adapt attention layers to MVS.

\paragraph{TransMVSNet vs. STTR} STTR~\cite{li2021revisiting} performs self- and cross-attention along intra- and inter-epipolar line to estimate stereo depth, where the context range of local features is only limited to their corresponding epipolar lines. Note that there does not exist line-to-line correspondence in MVS, and we thus utilize attention layers along whole flattened feature maps, to bring global context into feature matching over multiple views.

\subsection{Limitations}
Known limitations of TransMVSNet are listed below.
\begin{enumerate}
\setlength{\itemsep}{0pt}
\setlength{\parsep}{0pt}
\setlength{\parskip}{0pt}
    \item[-] Transformer slows down the speed of inference, as is shown in \cref{tab:ablation}.
    \item[-] Similar to other coarse-to-fine MVS networks, our method is sensitive to inference hyperparameters, \eg number of depth hypotheses, depth interval, and decay factor of depth interval.
\end{enumerate}

\section{Conclusion}
In this paper, we present a novel learning-based MVS network, termed as TransMVSNet, which aggregates global long-range context-aware information via Transformer. Specifically, TransMVSNet comprises an effective Feature Matching Transformer (FMT) module formulated with intra-attention and inter-attention, which focus on retrieving context-aware information within and across images respectively. Moreover, we design the Adpative Receptive Field (ARF) module and a transformed feature pathway to better facilitate the function of FMT.
By extensive experiments, we observe that TransMVSNet achieves state-of-the-art performance on DTU dataset, Tanks and Templates benchmark, and BlendedMVS dataset. Furthermore, we hope our attempt will provide some insight and motivate people to reconsider the fundamental roles of global context information in MVS matching framework.

{\small
\bibliographystyle{ieee_fullname}
\bibliography{egbib}
}

\appendix
\clearpage

\section{About One-to-many Matching Pattern}
\begin{figure}[ht]
    \centering
    \includegraphics[width=\linewidth]{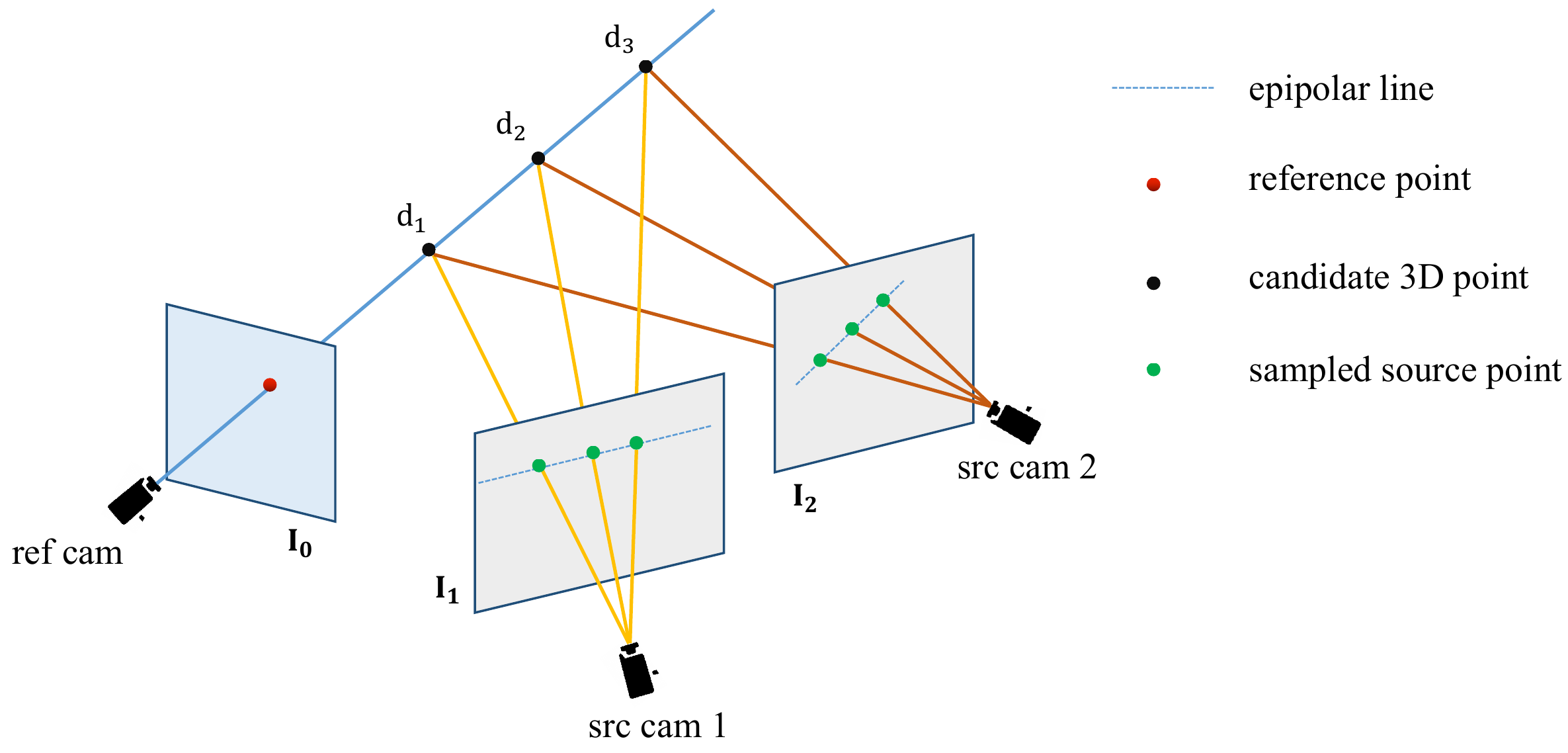}
    \caption{Illustration of the matching pattern of MVS.}
    \label{fig:one-to-many}
\end{figure}

As has been mentioned in the main paper, the nature of multi-view stereo (MVS) is a one-to-many matching task. We here further discuss the rationality of this analogy. As is illustrated in \cref{fig:one-to-many} where we consider a simple case that $N=3$,  for a pixel in the reference image, we attempt to find an optimal depth value $d$ among plane sweep depth hypotheses. These candidate 3D points all lie in an epipolar line of neighboring source images. In this way, the task of MVS becomes a typical one-to-many matching task where each pixel in $\mathbf{I}_0$ is supposed to find the best match among candidate source points in $\mathbf{I}_1$, and also, among candidate source points in $\mathbf{I}_2$.

\section{Ablation Study on Hyperparameters}
\subsection{Number of Views \& Input Resolution}
We conduct ablation study against the number of input views $N$ and input resolution $H\times W$ on DTU evaluation set~\cite{aanaes2016large}, and the results are listed in \cref{tab:NHW}. 
\begin{table}[ht]
    \footnotesize
    \centering
    \begin{tabular}{c|c|ccc}
     \hline
    $N$ & $H\times W$  & Acc. & Comp. & Overall \\
    \hline
    3 &  $864\times 1152$  & \textbf{0.310} & 0.323 & 0.316 \\
    5 &  $864\times 1152$  & 0.321 & \textbf{0.289} & \textbf{0.305} \\
    7 &  $864\times 1152$  & 0.327 & 0.295 & 0.311 \\
    9 &  $864\times 1152$  & 0.327 & 0.308 & 0.317 \\
    5 &  $512\times 640$  & 0.405 & 0.319 & 0.361 \\
     \hline
    \end{tabular}
    \caption{Ablation study on number of input views $N$ and image resolution $H\times W$ on DTU evaluation set~\cite{aanaes2016large} (\textbf{lower is better}).}
    \label{tab:NHW}
\end{table}


\subsection{Focusing Parameter}
We study the influence of training with different focusing parameter $\gamma$ of focal loss~\cite{lin2017focal} in \cref{tab:gamma}. Other inference settings are the same as the reported case in the main paper. By experiment results, $\gamma=0$ best fits the scene complexity of DTU dataset~\cite{aanaes2016large}, and obtains best Accuracy and Overall scores. As for BlendedMVS dataset~\cite{yao2020blendedmvs}, whose scenes are more diverse and complicated, $\gamma=2$ makes a big difference.
\begin{table}[ht]
    \footnotesize
    \centering
    \begin{tabular}{c|ccc|ccc}
    \hline
    \multirow{2}{*}{$\gamma$}& \multicolumn{3}{c|}{DTU} & \multicolumn{3}{c}{BlendedMVS}\\
     & Acc. & Comp. & Overall & EPE & $e_1$ & $e_3$ \\
    \hline
    0 & \textbf{0.321} & 0.289 & \textbf{0.305} & 0.80 & 9.79 & 4.40\\
    0.5 & 0.341 & 0.282 & 0.312 & 0.79 & 8.91 & 3.94\\
    1 &  0.342 & \textbf{0.279} & 0.310 & 0.78 & 8.80 & 3.86 \\
    2 &  0.345  & 0.282 & 0.314 & \textbf{0.73}  & \textbf{8.32} & \textbf{3.62}\\
    \hline
    \end{tabular}
    \caption{Ablation study on the value of $\gamma$ on DTU evaluation set~\cite{aanaes2016large} and BlendedMVS validation set~\cite{yao2020blendedmvs} (\textbf{lower is better}).}
    \label{tab:gamma}
\end{table}

\section{Ablation Study on FMT Design}
We further explore the architecture design of the Feature Matching Transformer (FMT) described in the main paper.

\subsection{Number of Attention Heads}
As is mentioned in the main paper, we apply multi-head attention where feature channels are split into $N_h$ groups, namely attention heads. We therefore study the influence of $N_h$, including model performance, memory consumption and inference time, in \cref{tab:numhead}. With $N_h$ increasing, there is no difference in memory consumption, but the inference time fluctuates possibly due to PyTorch's underlying implementation.

\begin{table}[ht]
    \footnotesize
    \centering
    \begin{tabular}{c|ccc|cc}
    \hline
    $N_h$ & Acc. & Comp. & Overall & Mem.(\textit{MB}) & Time(\textit{s})\\
    \hline
    1 & 0.322 & 0.296 & 0.309 & 3778 & \textbf{0.706} \\
    2 & 0.322 & 0.290 & 0.306 & 3778 & 1.024 \\
    4 & 0.322 & 0.289 & 0.306 & 3778 & 1.020 \\
    8 & \textbf{0.321} & \textbf{0.289} &\textbf{0.305} & 3778 & 0.996 \\
    16 & \textbf{0.321} & 0.290 & 0.306 & 3778 & 1.040 \\
    \hline
    \end{tabular}
    \caption{Ablation study on the number of attention heads $N_h$ on DTU evaluation set~\cite{aanaes2016large} (\textbf{lower is better}).}
    \label{tab:numhead}
\end{table}

\subsection{Number of Attention Blocks}
We adjust the number of attention blocks $N_a$ and \cref{tab:numblock} shows respective evaluation results, memory occupancy and inference time. As is demonstrated, $N_a=4$ achieves a balance between performance and efficiency. 
\begin{table}[ht]
    \footnotesize
    \centering
    \begin{tabular}{c|ccc|cc}
    \hline
    $N_a$ & Acc. & Comp. & Overall & Mem.(\textit{MB}) & Time(\textit{s}) \\
    \hline
    2 & 0.337 & \textbf{0.288} & 0.312 & \textbf{3760} & \textbf{0.815}\\
    4 & 0.321 & 0.289 & \textbf{0.305} & 3778 & 0.996 \\
    6 & \textbf{0.319} & 0.298 & 0.308 & 3792 & 1.223\\
    \hline
    \end{tabular}
    \caption{Ablation study on the number of attention blocks on DTU evaluation set~\cite{aanaes2016large} (\textbf{lower is better}).}
    \label{tab:numblock}
\end{table}

\subsection{Design of Attention Block}
\begin{figure}[ht]
  \centering
  \begin{subfigure}{\linewidth}
    \centering
    \framebox[\linewidth]{\includegraphics[width=0.75\linewidth]{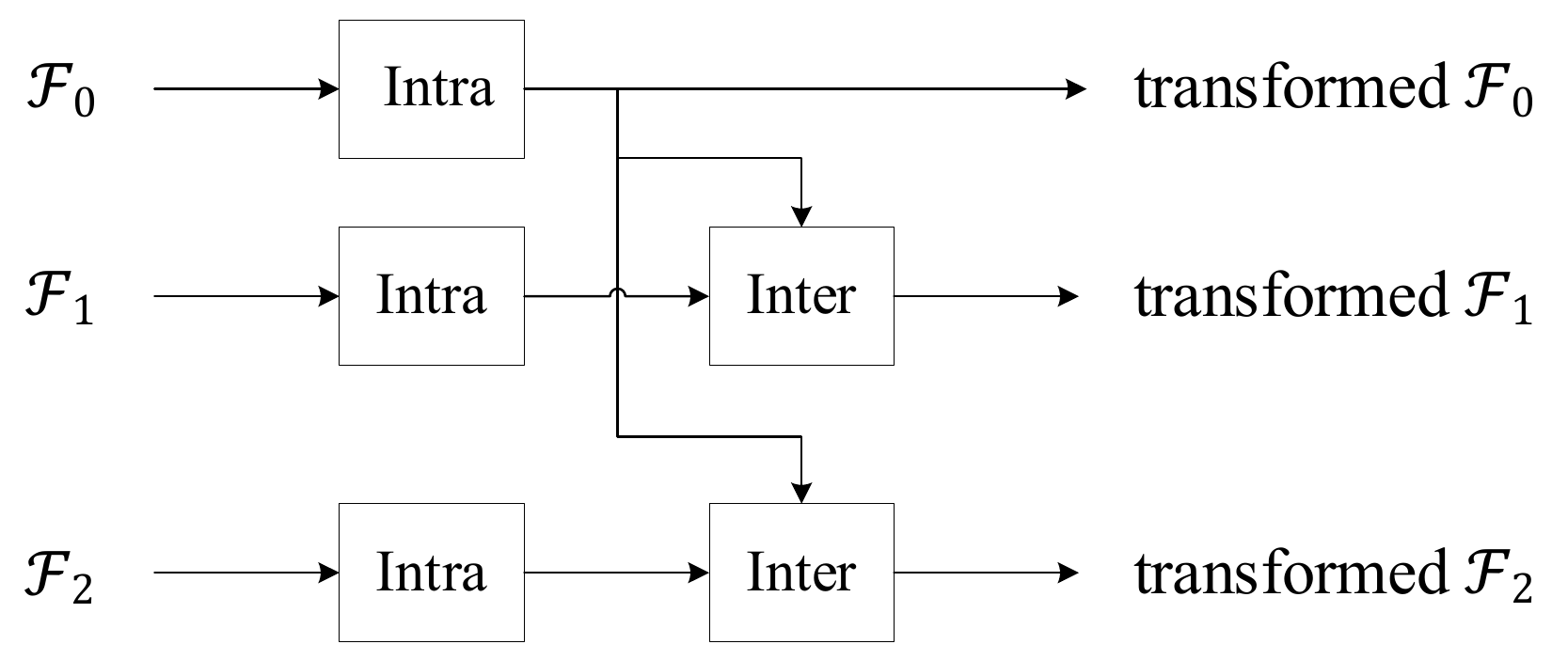}}
    \caption{}
    \label{fig:arch-a}
  \end{subfigure}
  \begin{subfigure}{\linewidth}
    \centering
    \framebox[\linewidth]{\includegraphics[width=0.99\linewidth]{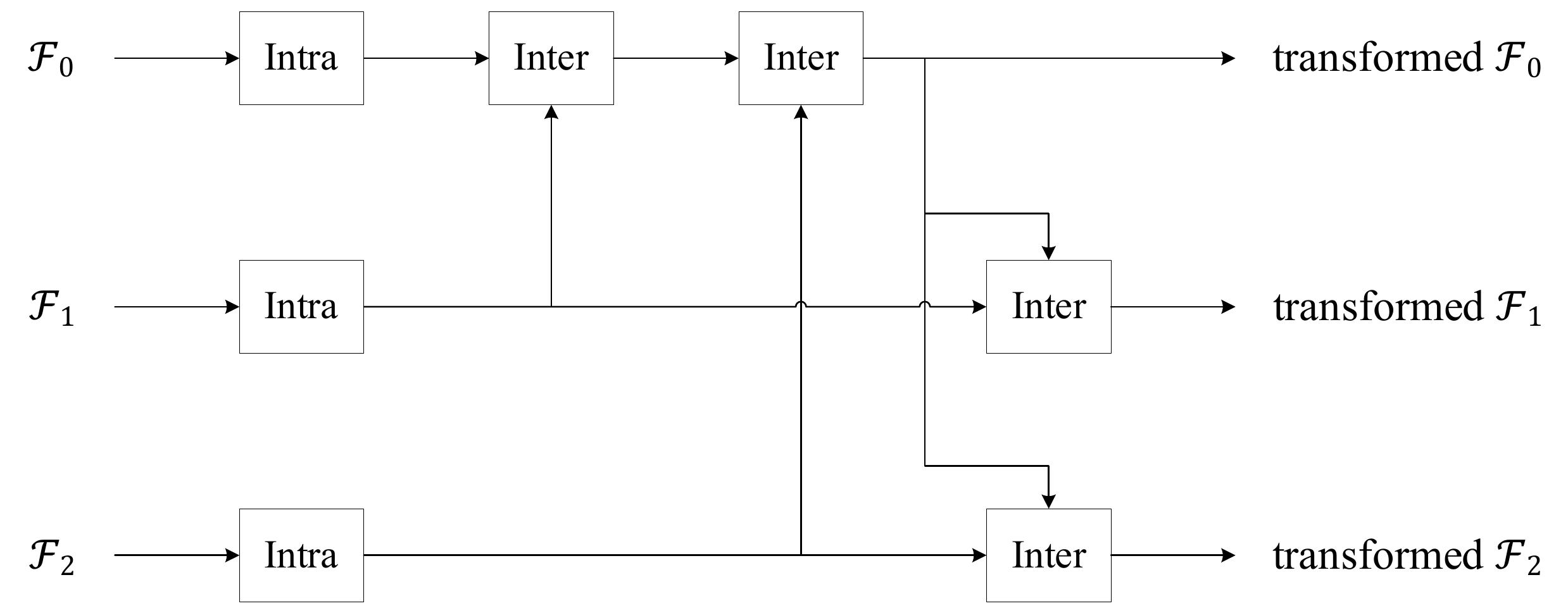}}
    \caption{}
    \label{fig:arch-b}
  \end{subfigure}
  \begin{subfigure}{\linewidth}
    \centering
    \framebox[\linewidth]{\includegraphics[width=0.99\linewidth]{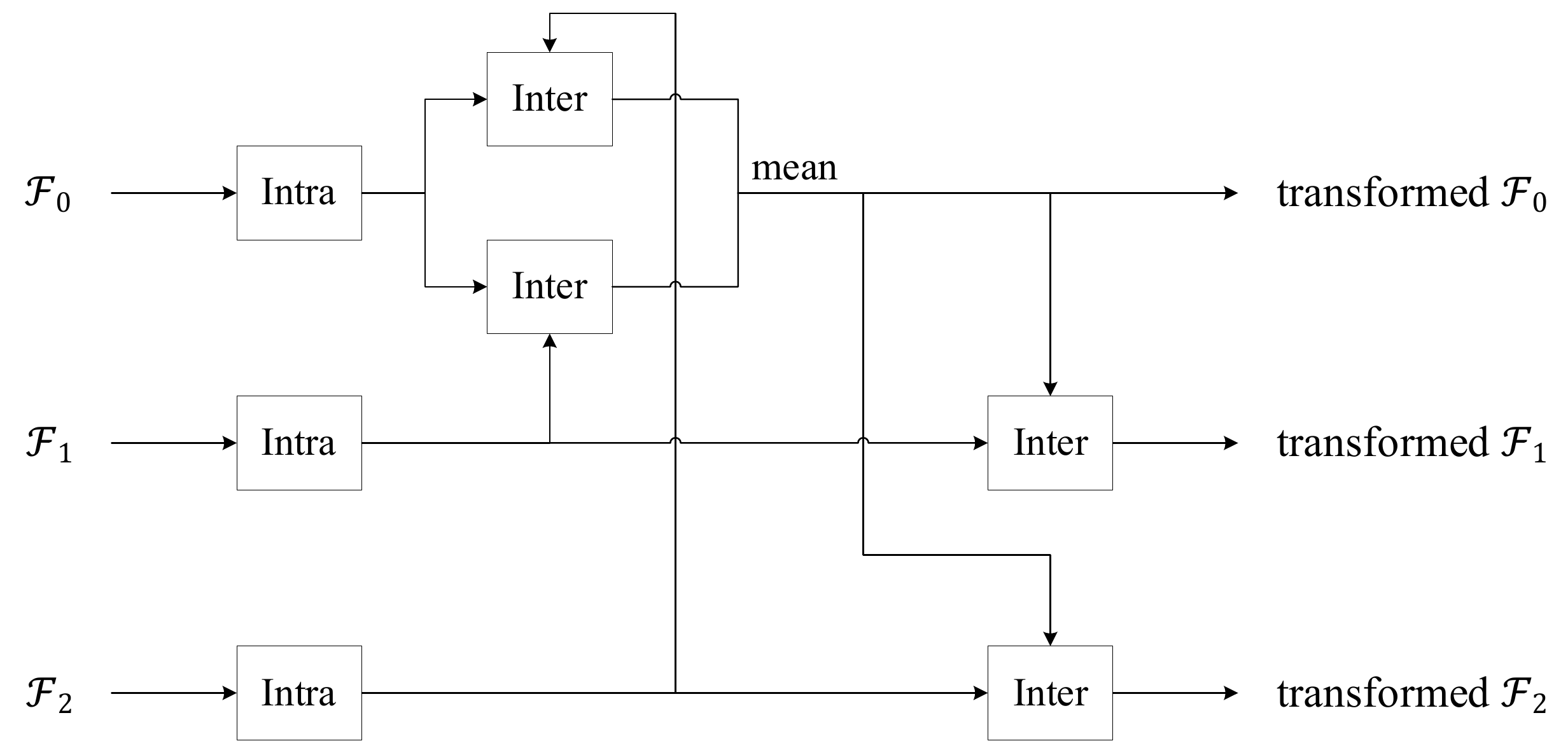}}
    \caption{}
    \label{fig:arch-c}
  \end{subfigure}
  \begin{subfigure}{\linewidth}
    \centering
    \framebox[\linewidth]{\includegraphics[width=0.72\linewidth]{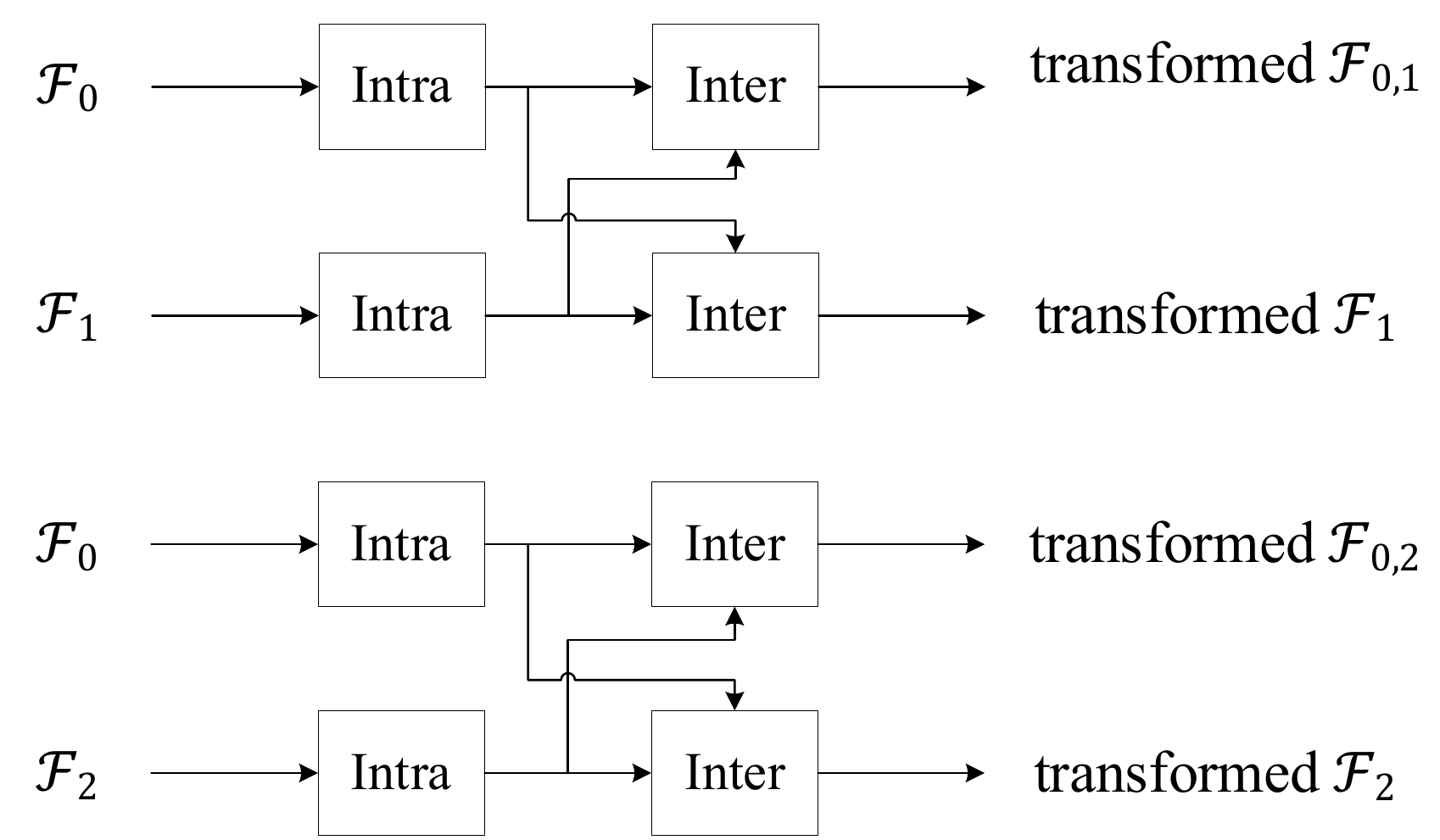}}
    \caption{}
    \label{fig:arch-d}
  \end{subfigure}
  \caption{Candidate designs of attention block. Note that there are in total $N-1$ source feature maps and we omit most of them for brevity.}
  \label{fig:arch_candidate}
\end{figure}

During the exploration, we study several potential architectures for attention block. We assume that intra-attention is always performed upon both the reference feature $\mathcal{F}_0$ and source features $\{\mathcal{F}_i\}_{i=1}^{N-1}$. Therefore the main differences lie in how to handle the reference feature $\mathcal{F}_0$ under inter-attention. We present all 4 possible designs covered as follows. The 4 possible candidate designs are illustrated in \cref{fig:arch_candidate}.

(a) Only reference-to-source inter-attention is performed upon $\mathcal{F}_0$, so $\mathcal{F}_0$ is only updated by intra-attention. \textbf{This is the final choice in TransMVSNet.}

(b) We sort $N-1$ source images with the same view selection protocol used in \cite{yao2018mvsnet} and perform source-to-reference inter-attention sequentially. Then the reference-to-source inter-attention is done.

(c) We duplicate $\mathcal{F}_0$ into $N-1$ identical copies so that source-to-reference inter-attention can be performed pair-wise in parallel. For the reference-to-source inter-attention, we average all $N-1$ transformed $\mathcal{F}_0$.

(d) $\mathcal{F}_0$ is duplicated at the very beginning of FMT so that each source and the reference feature form a pair throughout the whole FMT. Inter-attention operations of both directions, in this way, are performed at the same time and the final outputs of FMT are $N-1$ respectively transformed $\mathcal{F}_{0,i}(i=1,\ldots,N-1)$ and $N-1$ transformed source features.

To conclude, $\mathcal{F}_0$ in (b)(c)(d) is updated by source-to-reference inter-attention. (b) and (c) differ in the order of inter-attention: (b) does it sequentially while (c) is in parallel. (a)(b)(c) all follow a one-to-many matching pattern while (d) explicitly splits it into $N-1$ one-to-one problems. We also quantitatively study different designs in terms of both performance and costs. As is shown in \cref{tab:arch_candidate}, memory consumption of (a)(b)(c) is identical but (a) is more efficient in inference time. (d) occupies significantly more memory and takes more time than other candidates. As a result, (a) is both effective and efficient, verifying the intuition that under a one-to-many matching scenario, the matching target (the reference feature) should always be identical.

\begin{table}[ht]
    \footnotesize
    \centering
    \setlength\tabcolsep{1.8pt}
    \resizebox{\linewidth}{!}{
    \begin{tabular}{c|cc|ccc|cc}
    \hline
     & Matching & $\mathcal{F}_0$ Updated & \multirow{2}{*}{Acc.} & \multirow{2}{*}{Comp.} & \multirow{2}{*}{Overall} & Mem. & Time\\
     & Pattern & by Inter-att. & & & & (\textit{MB})& (\textit{s}) \\
    \hline
    (a) &  one-to-many & no & 0.321 & \textbf{0.289} & \textbf{0.305} & \textbf{3778} & \textbf{0.996} \\
    (b) &  one-to-many & yes &  \textbf{0.320} & 0.304 & 0.312 & \textbf{3778} & 1.197\\
    (c) &  one-to-many & yes & 0.332 & 0.294 & 0.313 & \textbf{3778} & 1.178 \\
    (d) &  one-to-one & yes & 0.339 & 0.292 & 0.316 & 4142 & 1.331 \\
    \hline
    \end{tabular}}
    \caption{Ablation study on the architecture design of attention blocks on DTU evaluation set~\cite{aanaes2016large} (\textbf{lower is better}).}
    \label{tab:arch_candidate}
\end{table}

\section{Visualized Attention}
\begin{figure*}[t]
  \centering
   \includegraphics[width=0.75\linewidth]{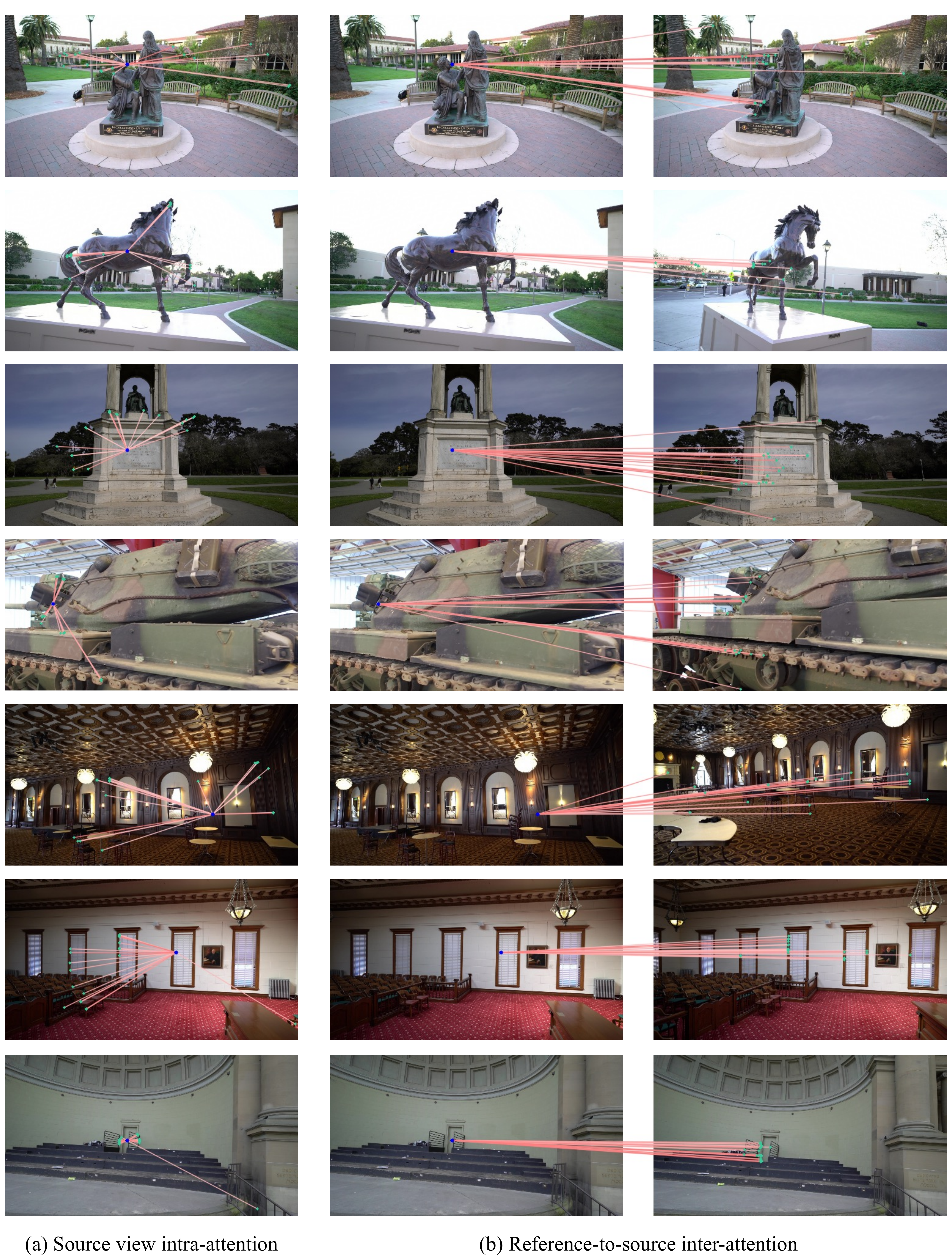}
   \caption{Visualization of intra- and inter-attention weights. From left to right, the first column (a) shows the intra-attention weights of a query point in the source image; the second and the third columns (b) show the inter-attention weights of the same query point in the source image with regard to its top-20 correspondences in the reference image.}
   \label{fig:supp_weight_map}
\end{figure*}

We visualize the weights of both intra- and inter-attention in \cref{fig:supp_weight_map}. For query points from challenging regions, \eg textureless or non-Lambertian surfaces, intra-attention seeks context information globally and inter-attention tends to match features across images. These two attention mechanisms are complementary and beneficial to robust depth estimation at challenging regions.


\section{Visualized Feature Map}
\begin{figure*}[t]
  \centering
   \includegraphics[width=0.95\linewidth]{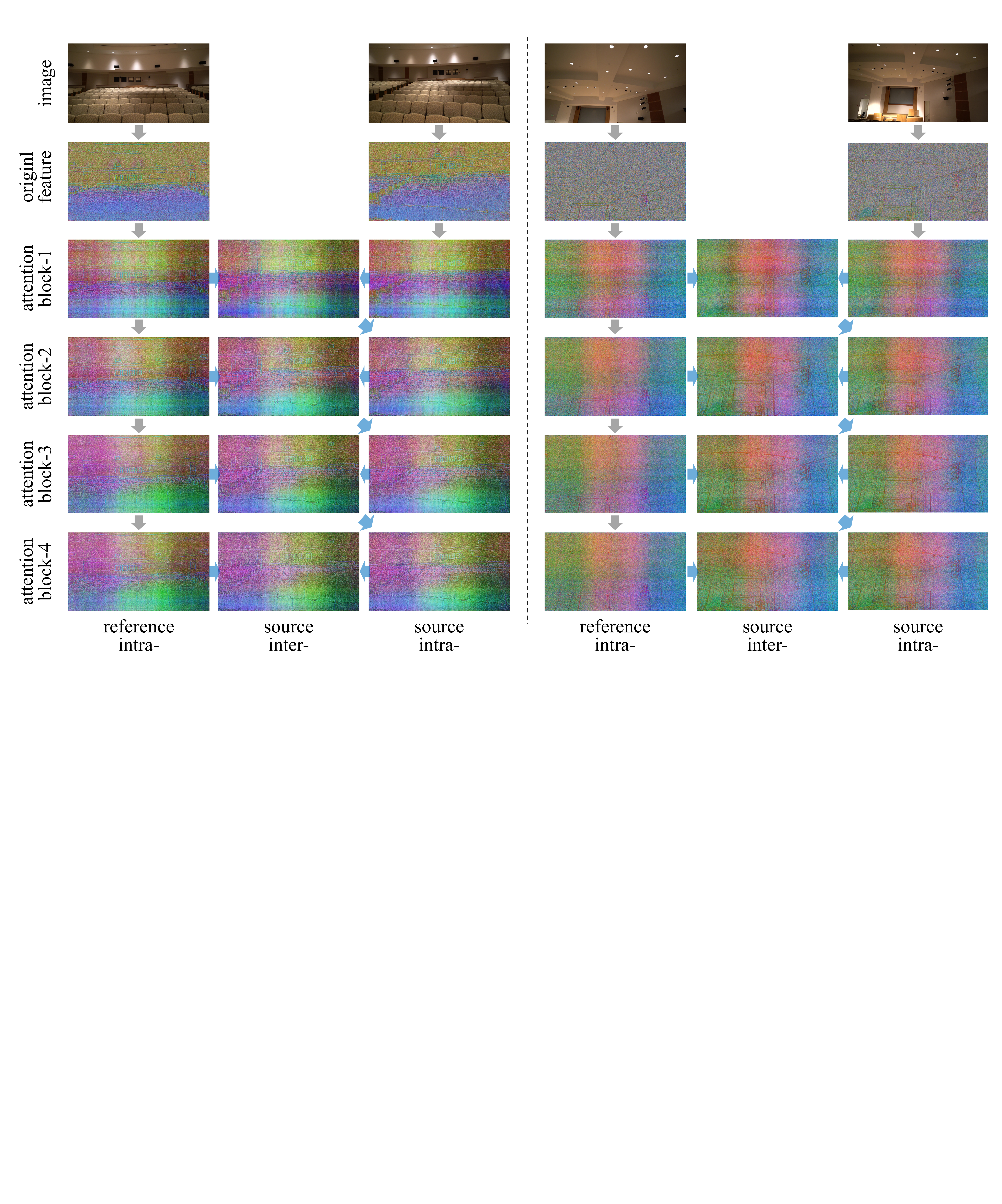}
   \caption{Evolution of transformed feature after each attention block. For better visualization, we apply PCA to reduce the number of feature channels to 3 and color the channels with RGB.}
   \label{fig:supp_feature}
\end{figure*}

To better illustrate the evolution of feature maps throughout FMT, we give another group of examples to demonstrate how FMT changes the feature in \cref{fig:supp_feature}. Before FMT, the extracted feature maps from FPN are not recognizable enough for robust feature matching at challenging areas since the feature representation is mostly local. After several intra- and inter-attention modules, more global position-dependent context information is encoded into the feature map, which benefits feature matching for textureless and non-Lambertian surfaces.


\section{More Point Cloud Results}

\begin{figure*}[t]
  \centering
   \includegraphics[width=\linewidth]{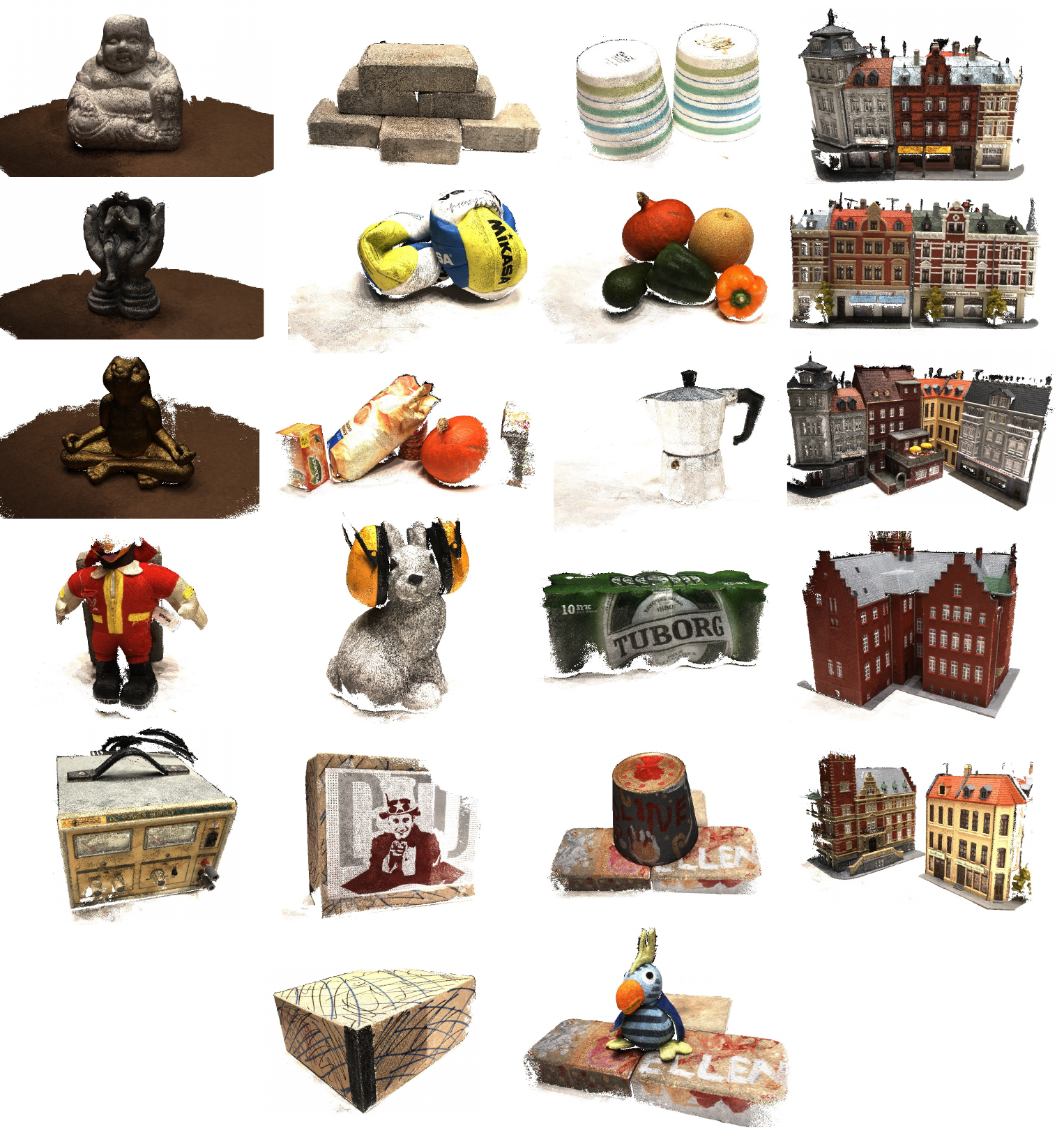}
   \caption{Point clouds of all 22 scans in DTU evaluation set~\cite{aanaes2016large} reconstructed by TransMVSNet.}
   \label{fig:dtu_points}
\end{figure*}

\begin{figure*}[t]
  \centering
   \includegraphics[width=\linewidth]{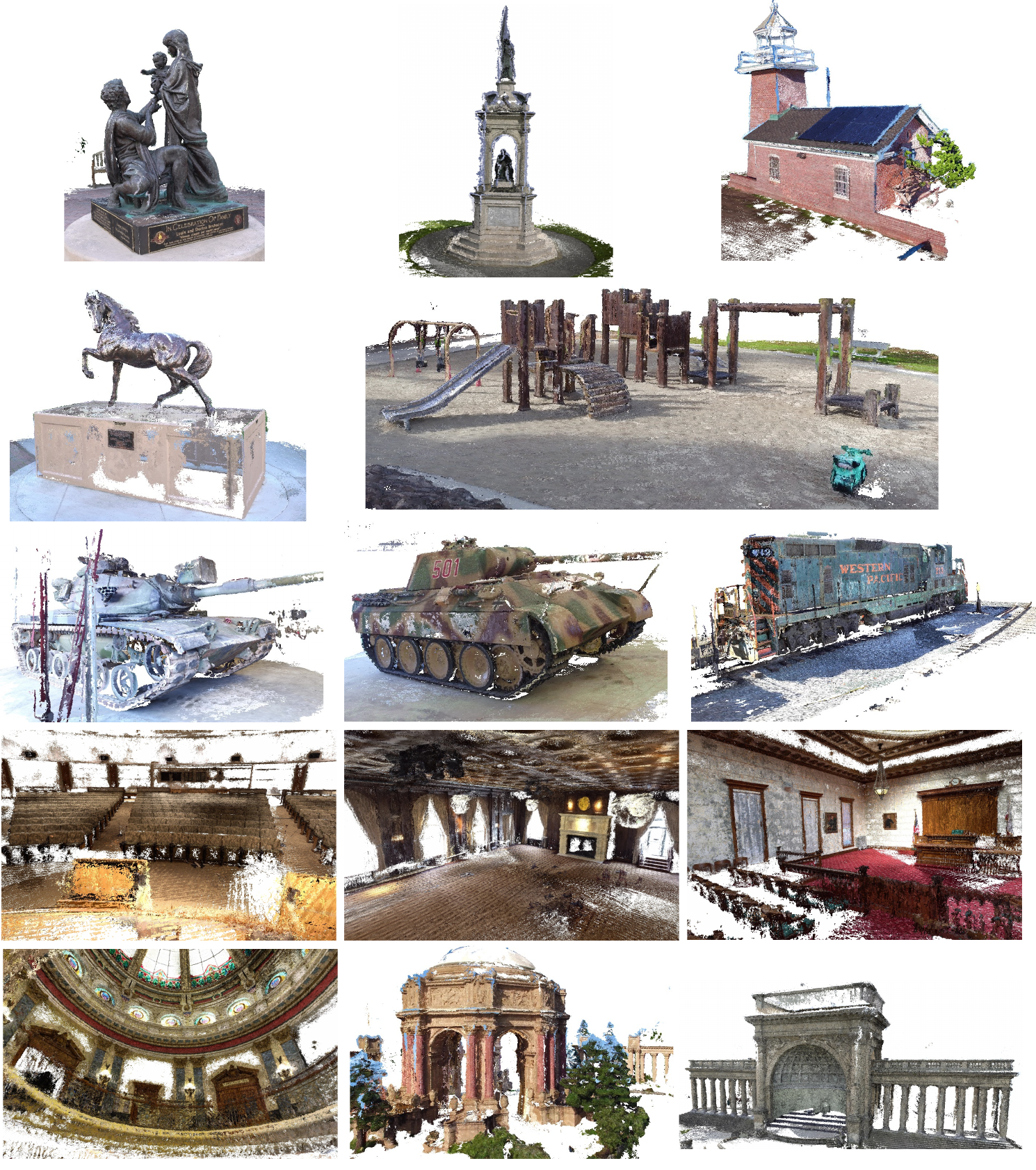}
   \caption{All point clouds of Tanks and Temples Benchmark~\cite{knapitsch2017tanks} (intermediate \& advanced) reconstructed by TransMVSNet.}
   \label{fig:tnt_points}
\end{figure*}

\begin{figure*}[t]
  \centering
   \includegraphics[width=0.75\linewidth]{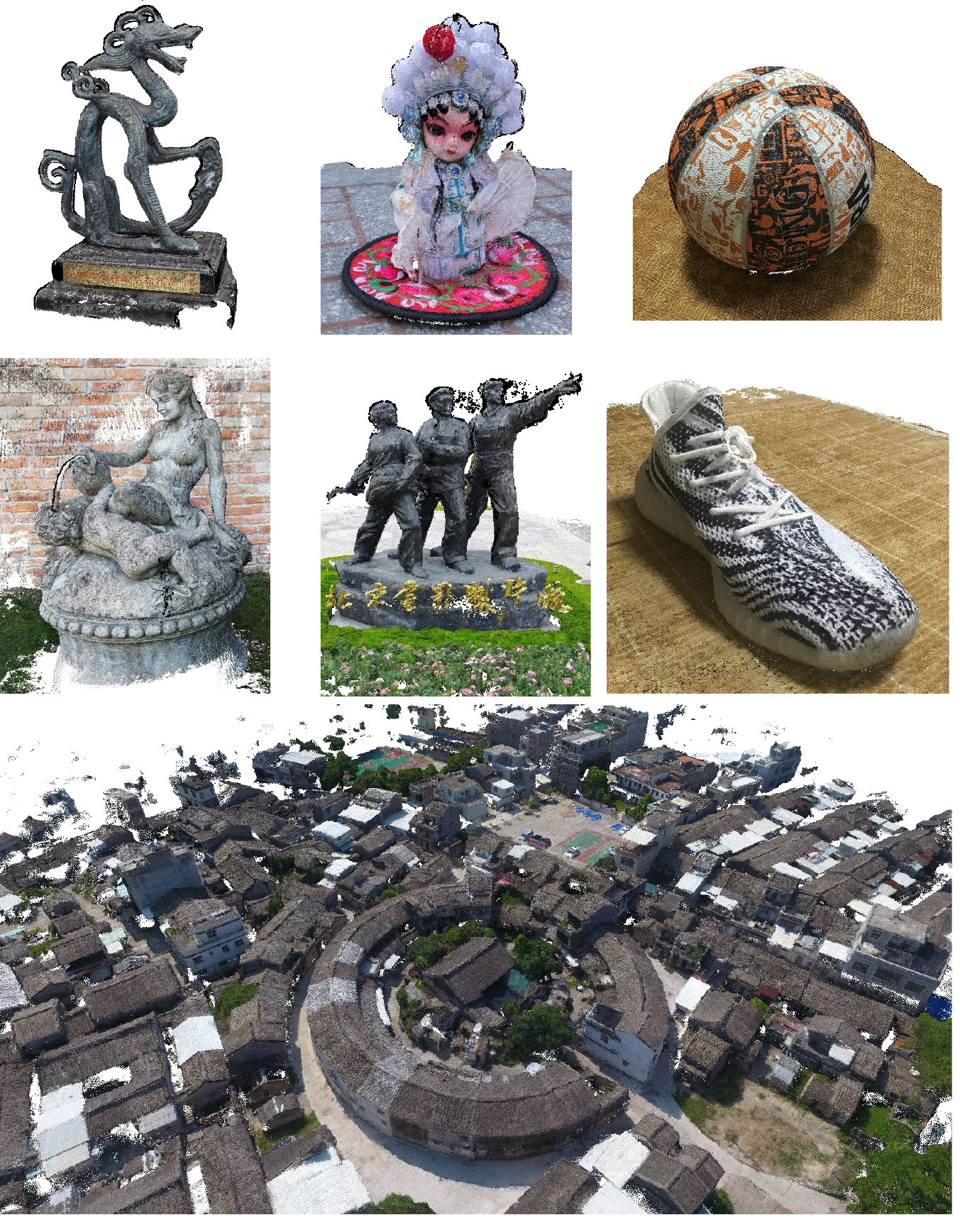}
   \caption{Point clouds of all 7 scenes in BlendedMVS validation set~\cite{yao2020blendedmvs} reconstructed by TransMVSNet.}
   \label{fig:blended_points}
\end{figure*}

We visualize all results of DTU evaluation set~\cite{aanaes2016large}, the intermediate and advanced set of Tanks and Temples benchmark~\cite{knapitsch2017tanks} and BlendedMVS validation set~\cite{yao2020blendedmvs} respectively in \cref{fig:dtu_points}, \cref{fig:tnt_points} and \cref{fig:blended_points}. Our TransMVSNet demonstrates its robustness and scalability on scenes with varying depth ranges.

\section{Use of Existing Assets}
The implementation of TransMVSNet is based on \href{https://github.com/alibaba/cascade-stereo}{CasMVSNet}~\cite{gu2020cascade}, who also heavily borrows code from \href{https://github.com/xy-guo/MVSNet_pytorch}{the PyTorch version of MVSNet}~\cite{yao2018mvsnet}.

Preprocessed images and camera parameters of both DTU dataset~\cite{aanaes2016large} and Tanks and Temples benchmark~\cite{knapitsch2017tanks} are from \href{https://github.com/YoYo000/MVSNet}{the official repository} of MVSNet~\cite{yao2018mvsnet} \& R-MVSNet~\cite{yao2019recurrent}, where COLMAP-SfM~\cite{schonberger2016structure} is adopted to obtain the camera calibration for Tanks and Temples.

{\small
\bibliographystyle{latex/ieee_fullname}
}

\end{document}